\documentclass{ieeeaccess}
\usepackage{cite}
\usepackage{amsmath,amssymb,amsfonts}
\usepackage{algorithmic}
\usepackage{graphicx}
\usepackage{textcomp}
\usepackage[numbers]{natbib}
\usepackage{graphicx}
\usepackage{float}
\usepackage{subcaption}
\usepackage{array}
\usepackage{makecell}
\usepackage{setspace}
\usepackage[utf8]{inputenc}
\usepackage{amsfonts}
\newcommand\scalemath[2]{\scalebox{#1}{\mbox{\ensuremath{\displaystyle #2}}}}
\usepackage[intoc]{nomencl} 
\usepackage{upgreek}
\usepackage{bm}
\usepackage{mathtools}   
\usepackage{amsfonts}
\usepackage[linewidth=1pt]{mdframed}
\usepackage[justification=centering]{caption}
\usepackage{hyperref}

\begin{document}
\doi{XX.XXXX/ACCESS.20XX.DOI}
\title{\LARGE Systematic Online Tuning of Multirotor UAVs for Accurate Trajectory Tracking Under Wind Disturbances and In-Flight Dynamics Changes }
\author{\small \uppercase{AbdulAziz Y. AlKayas}\authorrefmark{2,3,*}\href{https://orcid.org/0000-0002-3218-724X}{\includegraphics[scale=0.75]{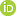}}, \uppercase{Mohamad Chehadeh}\authorrefmark{1,3,*}\href{https://orcid.org/0000-0002-9430-3349}{\includegraphics[scale=0.75]{orcid.png}}, \uppercase{Abdulla Ayyad}\authorrefmark{1,3}\href{https://orcid.org/0000-0002-3006-2320}{\includegraphics[scale=0.75]{orcid.png}}, and \uppercase{Yahya Zweiri}\authorrefmark{1,3}\href{https://orcid.org/0000-0003-4331-7254}{\includegraphics[scale=0.75]{orcid.png}}}
\address[1]{Aerospace Engineering Department, Khalifa University, Abu Dhabi, United Arab Emirates}
\address[2]{Mechanical Engineering Department, Khalifa University, Abu Dhabi, United Arab Emirates}
\address[3]{Khalifa University Center for Autonomous Robotic Systems (KUCARS), Khalifa University, Abu Dhabi, United Arab Emirates}
\tfootnote{Email addresses: 100052628@ku.ac.ae (AbdulAziz Y. AlKayas) ; mohamad.chehadeh@ku.ac.ae (Mohamad Chehadeh) ; abdulla.ayyad@ku.ac.ae (Abdulla Ayyad) ; yahya.zweiri@ku.ac.ae (Yahya Zweiri).}

\markboth
{Abdulaziz Alkayas \headeretal: Systematic Online Tuning of Multirotor UAVs for Accurate Trajectory Tracking}
{Abdulaziz Alkayas \headeretal: Systematic Online Tuning of Multirotor UAVs for Accurate Trajectory Tracking}

\corresp{*Corresponding authors, equally contributing.}

\begin{abstract}
The demand for accurate and fast trajectory tracking for multirotor Unmanned Aerial Vehicles (UAVs) have grown recently due to advances in UAV avionics technology and application domains. In many applications, the multirotor UAV is required to accurately perform aggressive maneuvers in challenging scenarios like the presence of external wind disturbances or in-flight payload changes. In this paper, we propose a systematic controller tuning approach based on identification results obtained by a recently developed Deep Neural Networks with the Modified Relay Feedback Test (DNN-MRFT) algorithm. We formulate a linear equivalent representation suitable for DNN-MRFT using feedback linearization. This representation enables the analytical investigation of different controller structures and tuning settings, and captures the non-linearity trends of the system. With this approach, the trade-off between performance and robustness in design was made possible which is convenient for the design of controllers of UAVs operating in uncertain environments. We demonstrate that our approach is adaptive and robust through a set of experiments, where accurate trajectory tracking is maintained despite significant changes to the UAV aerodynamic characteristics and the application of wind disturbance. Due to the model-based system design, it was possible to obtain low discrepancy between simulation and experimental results which is beneficial for potential use of the proposed approach for real-time model-based planning and fault detection tasks. We obtained RMSE of \(3.59 \; cm\) when tracking aggressive trajectories in the presence of strong wind, which is on par with state-of-the-art.
\end{abstract}

\begin{keywords}
Unmanned Aerial Vehicles, System Identification, Adaptive and Robust Control, Trajectory Tracking, PID Control, Machine Learning
\end{keywords}

\titlepgskip=-15pt

\maketitle

\section{Introduction}
\subsection{Motivation}
\PARstart{T}{rajectory} tracking problem for multirotor Unmanned Ae\-rial Vehicles (UAVs) has attracted significant attention from the robotics research community  in recent years. This is mainly due to the wide range of potential applications where accurate and precise trajectory tracking are needed. These applications include agriculture, entertainment, security, delivery, 3D Mapping etc. \cite{droneshows,Perez2019}. For example, precision agriculture using multirotor UAVs require accurate spatio-temporal tracking to efficiently spray the pesticides at the place and the time it's needed \cite{MOGILI2018502}. Another example is the need for accurate trajectory tracking of the multirotors in entertainment applications to execute the required trajectory while avoiding any attainable collateral damage.

A new UAV application domain is enabled by the recent advancements in on-board UAV navigation \cite{Tong2019,bry2015aggressive,mohta2018fast,murali2019perception}. When navigating cluttered environments, the UAV dynamically plans local trajectories. Following these trajectories in the presence of sensor uncertainty, external disturbances, and controller inaccuracies can be challenging. Moreover, some multirotor UAV applications would introduce system changes while in operation, for example installation of a specific payload to accomplish a certain task, such as a gripper or a camera \cite{LIERET2020424,compliant2019,Suarez2020,Suarez2018}, or even package for delivery tasks. These payloads change the dynamics such as the mass, moment of inertia and aerodynamic behavior, which result in trajectory tracking performance degradation or even instability in certain cases. As a result, high performance trajectory tracking for multirotors attracted the attention of many researchers in recent years. Additional enhancement for the tracking performance would lead to improved operational safety, and minimization of losses and accidents in missions carried in tight or crowded areas. In this work, we address the problem of accurate aggressive trajectory tracking in the presence of external wind disturbance, and in the case of in-flight changes to the aerodynamic properties of the UAV.

Estimating multirotor aerodynamic effects is a challenging problem that is widely studied in literature \cite{pounds2010modelling,hoffmann2007quadrotor,Faessler_2018,Chehadeh2019,torrente2021data}. Its complexity arises from the dependency on many system states; not only the nonlinear relation with the velocity, but also on the projected area (which depends on the attitude and velocity direction), induced drag due to the propulsion system and other complex phenomena \cite{bangura2017,pounds2010modelling}. Furthermore, it is much harder to achieve the desired performance at high velocities and accelerations because of the increased aerodynamic effects that are difficult to model and compensate for due to their complexity. Thus, aerodynamics accurate modeling of a multirotor is a research challenge, and proper knowledge of such dynamics is essential to guarantee high performance.
\subsection{Related Work}
Trajectory tracking problem for UAVs is an active research topic in the robotics community. Various algorithms and system components work in harmony to complete a trajectory tracking task. Trajectory tracking problem can be mainly split into two parts: feasible trajectory generation, and closed loop error minimization controller (for simplicity and compatibility with literature, we refer to it as trajectory tracking in this work), where this work is mainly concerned with the latter. A trajectory reference needs to be generated prior to tracking and the most adopted class of techniques is to use piece-wise polynomial trajectories \cite{VJKumar2011}. Tracking performance of such trajectories can be greatly enhanced when reference states are generated, which requires the generated trajectory to be continuous, and the system to be differentially flat. In some sense, generating reference states transforms a trajectory tracking problem into a state tracking problem, greatly enhancing the tracking performance as shown by the results reported in  \cite{VJKumar2011,Faessler_2018,tal2020}.

Trajectory tracking methods can be classified into two broad categories: feedforward and robust feedback methods. Feedforward methods rely on repeatable disturbance models to compensate for them through inversion. The pioneering work of \cite{Schoellig2012} used iterative learning to synthesize a drag description and its associated feedforward terms. A more recent work by \cite{Faessler_2018} built up on the useful property of differential flatness where the authors proved that a multirotor model with linear rotor drag is differentially flat. The flatness property is utilized to compute feedforward control terms to achieve accurate tracking of trajectories which has efficiency advantages in implementation and tuning compared to iterative learning methods. The computation of each of these feedforward terms depends on the rotor drag coefficients which are obtained through an optimization method and are specific for every trajectory to be followed. The results showed enhanced performance in Lemniscate and circular trajectories reaching velocities up to 5 m/s and the authors manually compensated for time delays in the system. A more recent work from the same group could generalize over all trajectory shapes \cite{torrente2021data}. The authors used a Model Predictive Controller (MPC) for multirotors with the aerodynamic effects modeled using Gaussian processes (GPs). By training using previously recorded flight data, the GPs can predict the error in acceleration due to aerodynamic drag given the velocity of the multirotor. The controller was able to achieve 70\% reduction in the tracking error for aggressive trajectories such as the figure-eight trajectory, reaching velocities up to 14 m/s and accelerations exceeding 4g. A drawback of this method is being computationally demanding such that a ground computer is required to perform MPC calculations. Another work by \cite{spitzer2019inverting} leveraged a learning based technique to produce adequate feedforward terms to perform accurate trajectory tracking in the presence of modeling and disturbance uncertainties. In a similar way to the other feedforward tuning methodologies, a regression algorithm learned a drag model based on flight data. All these methods \cite{Schoellig2012,Faessler_2018,torrente2021data,spitzer2019inverting} require extensive data collection and offline optimization to build up a drag or disturbance model, which limits the suitability for real-time adaptation. Also the optimized model can be biased towards the training data, and might under perform for unseen scenarios. For example, in \cite{Faessler_2018}, the drag model is refined for every considered trajectory shape. Trajectory tracking performance with feedforward methods can be significantly reduced due to the presence of external wind in the environment.

The other category of trajectory trackers utilize robust or adaptive controller structure and gains. One of the most common approaches for trajectory tracking is feedback linearization with differential flatness as suggested in \cite{VJKumar2011}. The recent work in \cite{tal2020} suggested an incremental nonlinear dynamic inversion (INDI) approach and was able to achieve high-performance trajectory tracking control scheme capable of successfully doing aggressive maneuvers (high accelerations and velocities) without the need for accurate modeling or knowledge about the aerodynamic parameters of the multirotor. They utilized the differential flatness property of the multirotor to generate the trajectory reference derivatives (velocity, acceleration, jerk and snap). To compensate for the inaccuracies in the model as well as the external disturbance due to aerodynamic drag, INDI control technique was used and arranged in a setting that requires high rate measurements of motor rotational speeds, and full states measurements. This is a drawback as such measurements are not available with the majority of the multirotor UAVs hardware currently deployed. The recent work of \cite{hamandi2020direct} utilized acceleration error scheme instead of the common feedback linearization based approaches. This approach requires a clean, lag-free acceleration measurement which was obtained by designing a novel regression based filter that removes accelerometer noise caused by propeller rotations (the accelerometer measurement comes from the IMU). This has a clear advantage over \cite{tal2020} as no additional sensors are required. In \cite{o2021meta}, the authors used model learning approach that adapts controller gains and showed improvement in tracking accuracy in the presence of external wind with speeds reaching \(6.2m/s\). However the achieved tracking performance was incomparable with the tracking performance for the wind free case. The presented experiments in \cite{o2021meta} were demonstrated in a scenario where the wind speed across all trajectory segments was constant. Other related work was based on combining different control and estimation methods to achieve the desired performance, like in \cite{NEKOUKAR2021104763} where the authors presented an adaptive fuzzy terminal sliding mode controller (AFTSMC) capable of tracking a predefined flight path under model uncertainty and external disturbances. Another work also based on the combination approach was investigated in \cite{GUO2020104560} where a multiple observers based anti-disturbance control (MOBADC) scheme was developed to enhance the tracking performance under wind and suspended payloads disturbances. One of the drawbacks was the need for partial information about the payload to maximize the rejection performance, but it showed good performance for the case of wind disturbance.

In all the surveyed work, there was no clear and systematic methodology for the automatic tuning of controller gains for trajectory tracking and adaptation to system changes; rat\-her, it depended on the human expertise or the extensive experimental data to achieve a satisfactory tracking performance. Based on the surveyed literature, it is not possible with the current state-of-the-art to obtain knowledge about the aerodynamic parameters in a form that is suitable for real-time controller synthesis. Also, current data-driven approaches lack predictability of the system response to unseen operating environments. Clearly, the current literature lacks a systematic simple and safe identification, and tuning methodology for accurate high speed trajectory tracking in the presence of external wind disturbances.

\subsection{Contribution}
In this paper, we propose a systematic approach for tuning and adapting controller parameters based on Deep Neural Networks and the Modified Relay Feedback Test (DNN-MRFT) for accurate high-speed trajectory tracking with disturbance attenuation capability. The proposed approach can adapt in real-time to changes in system dynamics that could happen during the mission, thus maintaining optimal trajectory tracking performance at all mission stages. DNN-MRFT ensures stability and sub-optimality performance bounds which we show to be negligible, thus optimality of performance can be claimed. DNN-MRFT runs in real-time on on-board computers and requires a few seconds to obtain optimal controller parameters for all UAV control loops.

We demonstrate the validity and efficiency of our approach by introducing significant change to the system dynamics during flight that would cause controller performance deterioration, which are four 12 cm wide and 40 cm long balloons fixed on each side of the multirotor. The new system is then identified and tuned using the DNN-MRFT approach regaining the state-of-the-art tracking performance for a figure-eight trajectory.

We also present the analysis of different tuning approach\-es for Proportional-Derivative (PD) and Proportional-Integral-Derivative (PID) controllers and their tracking performance under model parameter changes as well as external wind disturbances. Due to the model-based overall system design,low discrepancy between simulation and experimental results were obtained which proves the potential of using the proposed approach for real-time model-based planning and fault detection tasks. The obtained results show a trajectory tracking performance that is on par with the state-of-the-art control methods, and shows unique capability for attenuating external disturbances where the RMSE of tracking was \(3.59 \; cm\). A video of the experiments can be found in \cite{paper_video}.

\subsection{Structure of the Paper}
The paper is structured as follows. The model of multirotor UAV dynamics is presented in Section \ref{sec:dynamics_and_control}. The suggested feedback linearization approach and linearized model of the UAV with is presented in section \ref{sec:Control}. The methodology for identifying unknown system parameters is presented in Section \ref{sec:identification_dnn_mrft}. The analysis and tuning of the controller structures is presented in Section \ref{ControllerDesign}. Finally, simulation and experimental results are presented in Section \ref{sec:results}.
\section{Modeling Dynamics}
\label{sec:dynamics_and_control}
In this section, we present a nonlinear model of the multirotor UAVs that accounts for the actuator dynamics and digital delays in the system. We then present a linearized version of the nonlinear model that is suitable for identification and controller tuning. We choose a quadrotor as our simulation and experimentation platform, however this approach is directly extendable to any other symmetric multirotor UAVs.

\begin{figure}[h!]
    \includegraphics[width=\linewidth]{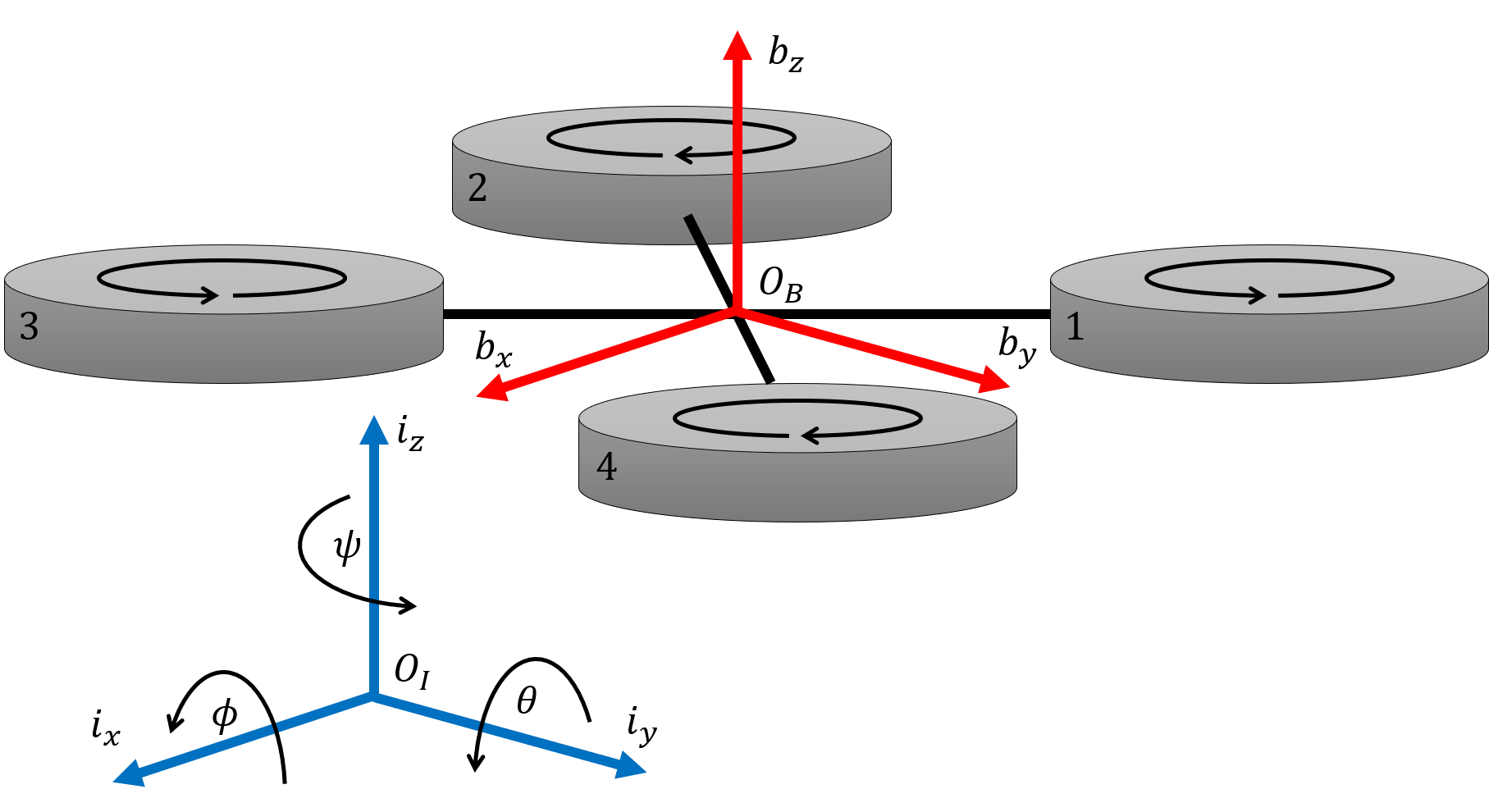}
        \caption{Schematic representation of the quadrotor and the coordinate systems used.}
        \label{fig:QCCoordSys}
\end{figure}
\subsection{Reference Frames and Conventions}
We define an Earth-fixed right-handed inertial frame $\mathcal{F}_I$ with the basis $\mathbf{i_z}$ pointing upwards opposite to gravity. It is convenient to express a vector with respect to a particular reference frame, e.g. the position vector ${}^I\mathbf{p} = [{}^Ip_x \; {}^Ip_y\; {}^Ip_z]^T$ is expressed in the inertial reference frame. Similarly, we define the body-fixed reference frame $\mathcal{F}_B$ to have a basis $\mathbf{b_z}$ that is parallel to the thrust force produced by the actuators, and is centered at the UAV center of gravity (CoG). Rotations around the inertial axes are given by the Euler angles \(\mathbf{\eta} = [ \phi \; \theta \; \psi]^T\) which describe the roll, pitch and yaw respectively. For convenience, we define the horizon frame $\mathcal{F}_H$ which has its origin coincident with the origin of $\mathcal{F}_I$. The reference frame $\mathcal{F}_H$ is yaw aligned with $\mathcal{F}_B$, while the pair \(\{\bm{h_z},\bm{i_z}\}\) is always coincident.
The orientation of the multirotor can be described by the rotation matrix ${}_B^IR$ which is a transformation defined in the SO(3) Lie group where the identity of the group is defined as ${}_B^IR{}_I^BR=\mathbf{I}$. The Lie algebra \(\bm{\mathfrak{so}}\bm{(3)}\) is defined around the identity element of the group with the rotational velocity vector \(\bm{\omega}=[p \; q \; r]^T\) \cite{hamel2006attitude}.

\subsection{Nonlinear Multirotor Body Dynamics}
The thrust force and pitching or rolling torques exerted by each propeller are described by:
\begin{equation}\label{eqn:rotorthrust}
\begin{split}
    f_i=k_f\Omega_i^2 , & \\    \uptau_{\phi,\theta_i}=k_fl_{\phi,\theta}\Omega^2_i . &  
\end{split}
\end{equation}




\begin{center}
 $i\in \{1,...,n_p\}$
\end{center}
 

Where \(i\) here denotes the rotor index, \(f\) is the thrust force, \(\uptau_{\phi,\theta}\) are rolling, pitching torques, \(k_f\) is the propeller thrust coefficient, \(l_{\phi,\theta}\) are the rolling, pitching moment arms, \(\Omega\) is the angular speed of the propellers. For the case of a quadrotor, we have the number of rotors \(n_p=4\). The yawing torque is different in nature than the rolling and pitching torques as it is caused by the reactive torque from the motors and it can be described by the following equation:

\begin{equation}
    \uptau_{\psi_{i}} = (-1)^{i+1} k_\uptau \Omega_i^2
\end{equation}

Where \(k_\uptau\) is the ropeller torque coefficient. Thus, the actuators forces and torques can be summarized in a matrix equation relating them to the angular speed $\Omega$ of each propeller:

\begin{equation}
\label{eq_motor_commands_to_force}
\begin{bmatrix}
\vspace{5pt} f_T \\ \vspace{5pt} \mathcal{\uptau}_\phi \\ \vspace{5pt} \mathcal{\uptau}_\theta \\ \vspace{5pt} \mathcal{\uptau}_\psi
\end{bmatrix} =
    \begin{bmatrix}
    \vspace{5pt} k_f & k_f & k_f & k_f \\
    \vspace{5pt} k_fl_{\phi} & -k_fl_{\phi} & -k_fl_{\phi} & k_fl_{\phi} \\
    \vspace{5pt} k_fl_{\theta} & k_fl_{\theta} & -k_fl_{\theta} & -k_fl_{\theta} \\
     k_\uptau & -k_\uptau & k_\uptau & -k_\uptau
    \end{bmatrix}
    \begin{bmatrix}
    \vspace{5pt} \Omega_1^2 \\ \vspace{5pt} \Omega_2^2 \\ \vspace{5pt} \Omega_3^2 \\ \Omega_4^2 
    \end{bmatrix}
\end{equation}



A multirotor is considered a rigid body in \(\mathbb{R}^3\), having 6 degrees-of-freedom (DOF) and subject to forces and torques in \(\mathbb{R}^3\). It is assumed that the weight of the multirotor is acting on CoG which is coincident with the body frame ${}^B\mathcal{F}$ origin, and assumed to be symmetric around all axes. Thus, the inertia matrix
is $\bm{J}= diag(J_{x},J_{y},J_{z})$. The governing equations for such bodies can be described by Newton-Euler equations described in the body frame as follows:

\begin{equation}\label{eqn:NEE}
\begin{bmatrix}
m\mathbf{I}_{3\times3} & 0_{3\times3}\\
0_{3\times3} & \bm{J}
\end{bmatrix}
\begin{bmatrix}
\bm{\dot{V}}\\
\bm{\dot{\omega}}
\end{bmatrix}
=\begin{bmatrix}
{}^B\bm{F}\\
{}^B\bm{\uptau}
\end{bmatrix}
\end{equation}


\begin{equation}\label{eqn:ForceB}
{}^B\bm{F} = f_T \bm{b_z} - {}_I^BRmg\bm{i_z} - \bm{\alpha}
\end{equation}


\begin{equation}\label{eqn:TorqueB}
    {}^B\bm{\uptau} = \begin{bmatrix}
\uptau_\phi \\ \uptau_\theta \\ \uptau_\psi
\end{bmatrix} - \bm{\lambda}
\end{equation}

where \(\bm{\dot{p}}={}_B^IR\bm{V}\), \(m\) is the mass, and $\bm{\alpha}$ and $\bm{\lambda}$ are arbitrary functions that describe the translational and rotational drag forces acting on the multirotors body respectively.

From equations (\ref{eqn:NEE}),(\ref{eqn:ForceB}) and (\ref{eqn:TorqueB})  and neglecting the cross-coupling dynamics due to their small and mitigated effect due to the closed loop performance, we achieve a simplified multirotor model:

\begin{equation}\label{eqn:QCmodel}
\begin{split}
    {}^I\ddot{p}_x &= \frac{1}{m}\left( (c_\psi s_\theta c_\phi + s_\psi s_\phi)f_T -\alpha_x({}^I\dot{p}_x,\eta,\Omega)  \right),  \\
    {}^I\ddot{p}_y &= \frac{1}{m} \left( (s_\psi s_\theta c_\phi - c_\psi s_\phi)f_T - \alpha_y({}^I\dot{p}_y,\eta,\Omega) \right)  ,  \\
    {}^I\ddot{p}_z &= \frac{1}{m} \left( c_\phi c_\theta f_T - g - \alpha_z({}^I\dot{p}_z,\eta,\Omega) \right) ,  \\
    \ddot{\phi} &= \frac{1}{J_x} \left( \uptau_\phi - \lambda_\phi(\omega,\Omega) \right), \\
    \ddot{\theta} &= \frac{1}{J_y} \left( \uptau_\theta - \lambda_\theta(\omega,\Omega) \right), \\
    \ddot{\psi} &= \frac{1}{J_z} \left( \uptau_\psi - \lambda_\psi(\omega,\Omega) \right). \\
\end{split}
\end{equation}

\subsection{Actuator Dynamics}
\label{sec_actuator_dynamics}
Brushless Direct Current (BLDC) motors is the common choice as an actuator for multirotors, and each motor requires a dedicated electronic speed controller (ESC). The ESC receives a thrust reference \(u_i\) and produces a thrust force \(f_i\). The relationship between \(u_i\) and \(f_i\) is usually quadratic as the ESC regulates the motor's rotational speed. We use ESCs that regulate the square of the motor's rotational speed, which results in a linear map between the ESC command and the generated thrust. This relationship can be approximated well by a First Order Plus Time Delay (FOPTD) transfer function as investigated in \cite{cheron2010,Chehadeh2019}:
\begin{equation}\label{eqn:actdyn}
    G_{prop}(s) = \frac{K_{eq}e^{-\tau_{act} s}}{T_{prop}s+1}
\end{equation}

Usually time delay is omitted from the models widely adopted in literature \cite{Faessler_2018,VJKumar2011}. From our experimentation and numerical simulations, we could not achieve acceptable tuning results when time delay is omitted \cite{Ayyad2020,Chehadeh2019}. Other researchers could minimize the effect of the omission of time delay by using high throughput sensor measurements, and the measurement of additional system states as motor's rotational speed and acceleration. Our approach alleviates such measurement requirements, thus it is applicable to numerous multirotor UAVs that exist in the market.

The commanded quantities (i.e. output of the controllers) are mapped to the individual motor commands by inverting the relation given in Eq. \eqref{eq_motor_commands_to_force}. Thus the commanded quantities and their physical counterparts (i.e. \(u_T\) with \(f_T\), \(u_\theta\) with \(\tau_\theta\), \(u_\phi\) with \(\tau_\phi\), and \(u_\psi\) with \(\tau_\psi\)) are always related by the actuator dynamics given in Eq. \eqref{eqn:actdyn}.




\begin{table*}
\centering
\begin{tabular}{||p{4cm}|p{10cm}|p{2cm}||}
\hline
Complexity Describtion & Addressing Method & Section\\ \hline
Nonlinear propulsion dynamics & ESCs that provide a linear map between the ESCs input commands and the generated thrusts are used. & Section \ref{sec_actuator_dynamics}  \\ \hline
Nonlinear kinematics due to gravity and under-actuation  & We use a feedback linearization law to obtain equivalent linear system. & Section \ref{sec:Control_fbl}  \\ \hline
Model parameters required for parametric tuning  & Model linearization around the hover operating point, and then performing identification using the DNN-MRFT approach which is suitable for linear SISO systems.& Section \ref{sec:identification_dnn_mrft} \\ \hline
Nonlinear drag dynamics  & A nominal linear drag term is obtained by DNN-MRFT identification. Controller parameters are then tuned to be robust against drag parameter uncertainty, thus achieving high performance for the whole operation envelope.  & Section \ref{sec:ControllerTuning}  \\ \hline
Motor saturation & Not addressed in the scope of this paper. But motor saturation is avoided in the trajectory generation stage to avoid operating in this regime. & -  \\ \hline
\end{tabular}
\caption{Summary of controller design considerations for various UAV model complexities.}
\label{table:controller_design_considerations}
\end{table*}

\section{Feedback Linearization and Model Linearization}\label{sec:Control}
The complexity of designing UAV controllers is mainly attributed to the coupling between the multiple control loops, nonlinear dynamics, and the uncertain dynamics of the UAV. We summarize these complexities and how we approached each of them in Table \ref{table:controller_design_considerations}. 

Sufficiently accurate state estimates are assumed to be available. But we account for the fact that they are delayed, for which the delay is found through identification. A kinematic based Kalman filter is used to provide smooth velocity and position estimates at the IMU update rate. The IMU is pre-calibrated and provide attitude and attitude rate estimates. Body accelerations, angular accelerations, and motor speeds are unknowns.

The strategy we follow to design the controller structure is to perform feedback linearization to transform the nonlinear system into an equivalent linear system. This linearization is valid as long as the motors are not saturated. The equivalent linear system is then utilized for identification, as would be discussed in Section \ref{sec:identification_dnn_mrft}. The detailed linear system models used gave us the freedom to design and tune controllers that perform well experimentally, with clear guidelines for the trade-off between performance and robustness.

\subsection{Feedback Linearization}\label{sec:Control_fbl}
We use a feedback linearization approach similar to the approaches suggested in \cite{VJKumar2011,Faessler_2018}. The position controllers provide the control desired acceleration vector \(\bm{{}^I\ddot{p}}^{d}\) as follows:
\begin{equation}
\label{eq:outer_loop_controller}
\begin{split}
    \bm{{}^I\ddot{p}}^{d}={}_H^IR\big(\bm{K}_{p}\odot{}^H({}^I\bm{p}^{ref}-{}^I\bm{p})+\bm{K}_{v}\odot{}^H({}^I\bm{\dot{p}}^{ref}-{}^I\bm{\dot{p}})\\
    +\bm{K}_{i}\odot{}^H({}^I\bm{\bar{e}_\bm{p}})\big)+{}^I\bm{\ddot{p}}^{ref}
\end{split}
\end{equation}
where \(\odot\) is the Hadamard product, \(\bm{\bar{e}_{\bm{p}}}\) is an augmented state as given in:
\begin{equation*}
   \bm{\bar{e}_{\bm{p}}}= \int_0^t{({}^I\bm{p}^{ref}-{}^I\bm{p})}dt\;,
\end{equation*}
and reference position and its derivatives are denoted by the \(ref\) superscript. The vectors \(\bm{K}_{p}\), \(\bm{K}_{v}\), and \(\bm{K}_{i}\) are fixed controller gains. 

The commanded acceleration output, \({}^I\bm{\ddot{p}}^c\), that is used to calculate motor commands and attitude loops references is given by:
\begin{equation}
    {}^I\bm{\ddot{p}}^c = {}^I\bm{\ddot{p}}^{d} + a_g\bm{i_z}
\end{equation}

where \(a_g\) corresponds to the estimated acceleration due to gravity (we assume it to be \(9.79\,m/s^2\). For convenience, we then define a temporal reference frame ${}^{C}\mathcal{F}$ that represents the commanded attitude with the basis \(\mathbf{c_z}\) aligned with ${}^I\bm{\ddot{p}}^c$ as follows:

\begin{equation}
\label{}
    \bm{c_z} = \frac{{}^I\bm{\ddot{p}}^{d}}{\| {}^I\bm{\ddot{p}}^{d} \|}
\end{equation}

\begin{equation}
    \bm{c_y} = \frac{\bm{c_z} \times [\cos(\psi_{ref}) \; \sin(\psi_{ref}) \; 0]^T}{\| \bm{c_z} \times [\cos(\psi_{ref}) \; \sin(\psi_{ref}) \; 0]^T \|}
\end{equation}

\begin{equation}
    \bm{c_x} = \bm{c_y} \times \bm{c_z}
\end{equation}
Then the rotation matrix \({}_I^{C}R\) is constructed from the basis \([\bm{c_x}\; \bm{c_y}\; \bm{c_z}]\) which can be used to find the rotation vector constituting current orientation errors as follows:

\begin{equation}
    \begin{bmatrix}
    e_x \\ e_y \\ e_z
    \end{bmatrix} = \bm{\varepsilon}({}_I^{C}R \;\;{}_B^IR)
\end{equation}
where $\bm{\varepsilon}$ is a function that converts the rotation matrix into the corresponding rotation vector representation of the attitude based on the inverse of the Rodrigues' rotation formula, which leads to the axis-angle representation given by:
\begin{align}
\begin{aligned}
    \label{eq:rodrigues_formula}
\delta=&\arccos{(\frac{\text{Tr}(R)-1}{2})} \\
\bm{e}=&\frac{1}{2sin(\delta)}\begin{bmatrix}
    R_{32}-R_{23} \\ R_{13}-R_{31} \\ R_{21}-R_{12}
    \end{bmatrix}\\
\bm{\delta}=& \delta \bm{e}
\end{aligned}
\end{align}

where \(R_{ij}\) corresponds to the rotation matrix element at the \(i^{th}\) row and the \(j^{th}\) column. The singularities in Eq. \eqref{eq:rodrigues_formula} due to small rotation angles or large rotation angles close to \(\pi\) are handled properly.

Finally, we can calculate the collective thrust command $u_T$ by projecting the commanded thrust vector to \(\bm{c_z}\):
\begin{equation}
\label{eq:uz_feedback_linearization}
    u_T = k_b{}^I\bm{\ddot{p}}^c \cdot \bm{c_z}
\end{equation}
where \(k_b\) is a dimensionless constant that maps the commanded acceleration to ESC command. Assuming a properly calibrated ESC with linearly proportional thrust response, \(k_b\) is found to be the hover thrust \(u_{T_0}\) divided by \(a_g\).

\subsection{Linearized Attitude and Altitude Dynamics}
\begin{figure*}
    \centering
    \includegraphics[width=\textwidth]{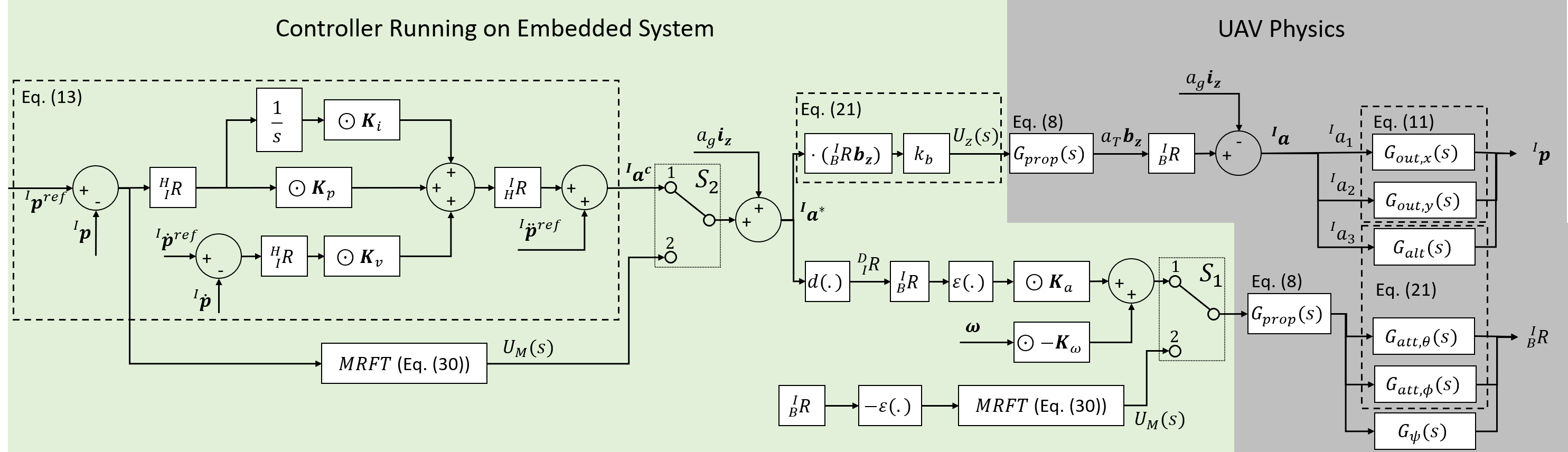}
        \caption{Controller and identification structure and their interface with the physical model. The switches \(S_1\) and \(S_2\) are set at position 1 in the control phase, and are set at position 2 during the identification phase.}
        \label{fig:controller_struct}
\end{figure*}


Near hover multirotor UAV operation, both the attitude and altitude dynamics can be approximated by a first order system plus integrator \cite{pounds2010modelling}:
\begin{equation}\label{eqn:attalttdyn}
    G_{att,alt}(s) = \frac{K_{eq}}{s(T_1s+1)}
\end{equation}
Such model accounts for the fact that drag dynamics cannot be neglected \cite{pounds2010modelling}. The input to this model is torque for the attitude dynamics case, and force for the altitude dynamics case. And the output is the attitude angle (i.e. either roll or pitch) or the altitude. The total loop dynamics that would include the actuator dynamics are given by cascading both transfer functions in \eqref{eqn:attalttdyn} and \eqref{eqn:actdyn} to obtain the following second order plus integrator with time delay (SOIPTD) model:

\begin{equation}\label{eqn:innerloopdyn}
    G_{in}(s) = \frac{K_{eq}e^{-\tau_{in} s}}{s(T_{prop}s+1)(T_1s+1)}
\end{equation}
The time delay parameter accounts for the delays in the forward and feedback paths. This model provides a mapping between the ESC commands as inputs, to the measured attitude or altitude. The adequacy of this model in describing UAV attitude and altitude dynamics has been proven experimentally in \cite{Chehadeh2019,Ayyad2020,ayyad2020multirotors}.


\subsection{Linearized Lateral Motion Dynamics}
As an underactuated system, the multirotor cannot achieve lateral (i.e. in the direction of \(\bm{i_x}\) or \(\bm{i_y}\)) motion through direct actuation commands, but instead it is achieved through changing the attitude of the multirotor in the direction of motion. Because of underactuation, we refer to these dynamics as \emph{outer} dynamics, whereas we refer to attitude and altitude dynamics as \emph{inner} dynamics, as they command the ESCs directly. A linearized model for such outer dynamics was investigated in \cite{ayyad2020multirotors}:
\begin{equation}\label{eqn:sidemotion}
    G_{out}(s)  = \frac{K_{eq}e^{-\tau_{out}s}}{s(T_2s+1)}
\end{equation}
Which provides a relation between the tilt angle and the lateral position. Drag on the quadrotor body is assumed to be linear, and it is associated with the parameter \(T_2\). The feedback linearization controller presented in Section \ref{sec:Control_fbl} ensures this linear relationship holds as long as the motors are not saturated. It is therefore possible to cascade the dynamics in equations \eqref{eqn:actdyn}, \eqref{eqn:innerloopdyn} and \eqref{eqn:sidemotion} to obtain the overall lateral motion dynamics, mapping ESC commands to lateral position: 
\begin{equation}\label{eqn:outerloopdyn}
    G_{lat}(s) = \frac{K_{eq}e^{-(\tau_{in}+\tau_{out}) s}}{s^2(T_{prop}s+1)(T_1s+1)(T_2s+1)}
\end{equation}
The time delay parameter accounts for the delays in the forward and feedback paths. The mapping between the angle and the thrust is accounted for in the gain \(K_{eq}\). We have now two sets (one set for each lateral axis) of rotational and translational drag terms modelled by the time constants \(T_1\) and \(T_2\) respectively.

\subsection{Trajectory Generation}\label{TrajGen}
In our work, we choose figure-eight (i.e. Lemniscate of Gerono) trajectory with fixed altitude as our testing and benchmarking trajectory. It can be described by the following equations:

\begin{equation}\label{eqn:TrajGen}
\begin{split}
        p^{ref}_y(t) & = r\cos (\sigma(t)) \\
        p^{ref}_x(t) & = \frac{r\sin(2\sigma(t))}{2} \\
        \sigma(t) & = \sum_{i=0}^nb_it^i
    \end{split}
\end{equation}

Where $\sigma(t)$ is a polynomial of $n^{th}$ order, $b_i$ is a polynomial coefficient, and $r$ is the trajectory radius. We follow the work of \cite{VJKumar2011} in optimizing $\sigma(t)$ in order to minimize the integral of the square of a chosen trajectory derivative (in our case we minimize the snap) over its period while enforcing a set of initial and terminal constraints. We have used the derivative of the generated trajectory, $\dot{p}^{ref}_x$ and $\dot{p}^{ref}_y$ as shown in Equation (\ref{eq:outer_loop_controller}) to enhance the tracking performance. We also used the second derivative $\ddot{p}^{ref}_x$ and $\ddot{p}^{ref}_y$ as an additional feedforward input to $\bm{{}^I\ddot{p}}^d$ as shown in Equation (\ref{eq:outer_loop_controller}). Higher derivatives of the reference signal beyond the second were not included, although it would enhance the tracking performance, as we needed knowledge about the generated thrust rate and the angular speed of the propellers which are not measured quantities in our setup.

The overall linearized and decoupled closed-loop altitude system dynamics including the feedforward reference signals provided by the trajectory generator is given by:

\begin{equation}
\label{eq_closed_loop_alt}
    \frac{Z(s)}{R_z(s)}=\frac{G_{in}(C_{in}+s^2)}{1+G_{in}C_{in}}
\end{equation}
and for the closed-loop lateral dynamics:
\begin{equation}
\label{eq_closed_loop_lat}
    \frac{X(s)}{R_x(s)}=\frac{G_{in,cl}G_{out}(C_{out}+s^2)}{1+G_{in,cl}G_{out}C_{out}}
\end{equation}
where \(C_{in}\) is the in loop controller, \(G_{in,cl}\) is the closed loop inner dynamics (i.e. with the inner loop controller), and \(C_{out}\) is the outer loop feedback controller. The closed-loop dynamics maps a reference signal to its corresponding measured quantity, e.g. \(R_z(s)\) is the reference altitude and \(Z(s)\) is the measured altitude. The terms \(G_{in}s^2\) in the numerator of Eq. \eqref{eq_closed_loop_alt}, and  \(G_{in,cl}G_{out}s^2\) in the numerator of in Eq. \eqref{eq_closed_loop_lat}, which are due to the use of the feedforward reference signal, reduce the relative degree of the systems, which result in a faster response of these systems.

\section{Identification with DNN-MRFT}
\label{sec:identification_dnn_mrft}
In this section we discuss the DNN-MRFT identification approach that is used to obtain the linear model parameters \cite{Ayyad2020,ayyad2020multirotors}. DNN-MRFT performs identification in real-time within seconds and guarantees near optimal performance. These features were demonstrated by the experiments in \cite{ayyad2020multirotors}, where UAVs of different sizes were able to perform full identification from take-off, without prior knowledge of dynamics.

\subsection{Generating Periodic Motion}
The idea of the identification is to excite a periodic motion in the system that reveals the unknown system dynamics. Measured system output is then fed to a deep neural network (DNN) which classifies the unknown process parameters and provides near optimal tuning. The periodic motion is excited using the modified relay feedback test (MRFT) which is given by \cite{Boiko2013}:

\begin{equation}\label{eq_mrft_algorithm}
\scalemath{0.9}{u_M(t)}=\scalemath{0.75}{
\left\{
\begin{array}[r]{l l}
h\; &:\; e(t) \geq b_1\; \lor\; (e(t) > -b_2 \;\land\; u_M(t-) = \;\;\, h)\\
-h\; &:\; e(t) \leq -b_2 \;\lor\; (e(t) < b_1 \;\land\; u_M(t-) = -h)
\end{array}
\right.}
\end{equation}
where \(b_1=-\beta e_{min}\) and \(b_2=\beta e_{max}\). \(e_{max}>0\) and \(e_{min}<0\) are respectively the last maximum and minimum values of the error signal after crossing the zero level; and \(u_M(t-)=lim_{\epsilon\rightarrow0^+ }u_M(t-\epsilon)\) is the last control output. Initially, the maximum and minimum error values are set as \(e_{max}=e_{min}=0\). \(\beta\) is a tunable parameter that defines the phase of the excited oscillations. Based on the describing function (DF) method, it could be shown that the MRFT achieves oscillations at the phase angle defined by the parameter \(\beta\) by satisfying the harmonic balance equation \cite{atherton1975}:
\begin{equation}\label{eq_hb}
N_d(a_0)G(j\Omega_0)=-1
\end{equation}
The DF of the MRFT is presented in \cite{Boiko2013} as:
\begin{equation}\label{eq_mrft_df}
N_d(a_0)=\frac{4h}{\pi a_0}(\sqrt{1-\beta^{2}}-j\beta)
\end{equation}

The MRFT is triggered when identification needs to be performed. Figure \ref{fig:controller_struct} shows the placement of MRFT within the control structure. When the identification of the roll, pitch, or the altitude dynamics is needed, switch \(S_1\) position is changed to 2 and the associated outer loop is temporarily disabled. Each of the loops need to be identified independently. Once a steady-state limit cycle is produced, the switch \(S_1\) immediately returns to position 1 and the outer loop control is resumed for normal control operation using the newly tuned controllers based on the identification results. The generation of limit cycles for the outer loop dynamics requires the switch \(S_2\) to change to position 2 while \(S_1\) remains at position 1. Similar to the inner loops identification, once a steady-state limit cycle is produced, the switch \(S_2\) immediately returns to position 1 for the resumption of normal control operation using the newly acquired controller parameters.

Stability of the periodic motion for the considered multirotor UAV dynamics was proven in \cite{ayyad2020multirotors}. It was shown in \cite{Ayyad2020} that the DNN-MRFT approach achieves identification results that provide better controller performance in shorter time and with much smaller computational requirements compared with other identification and tuning methods, like the prediction error method and non-parametric tuning of PID controllers.

\begin{figure}[t]
\begin{center}
\includegraphics[width=\linewidth]{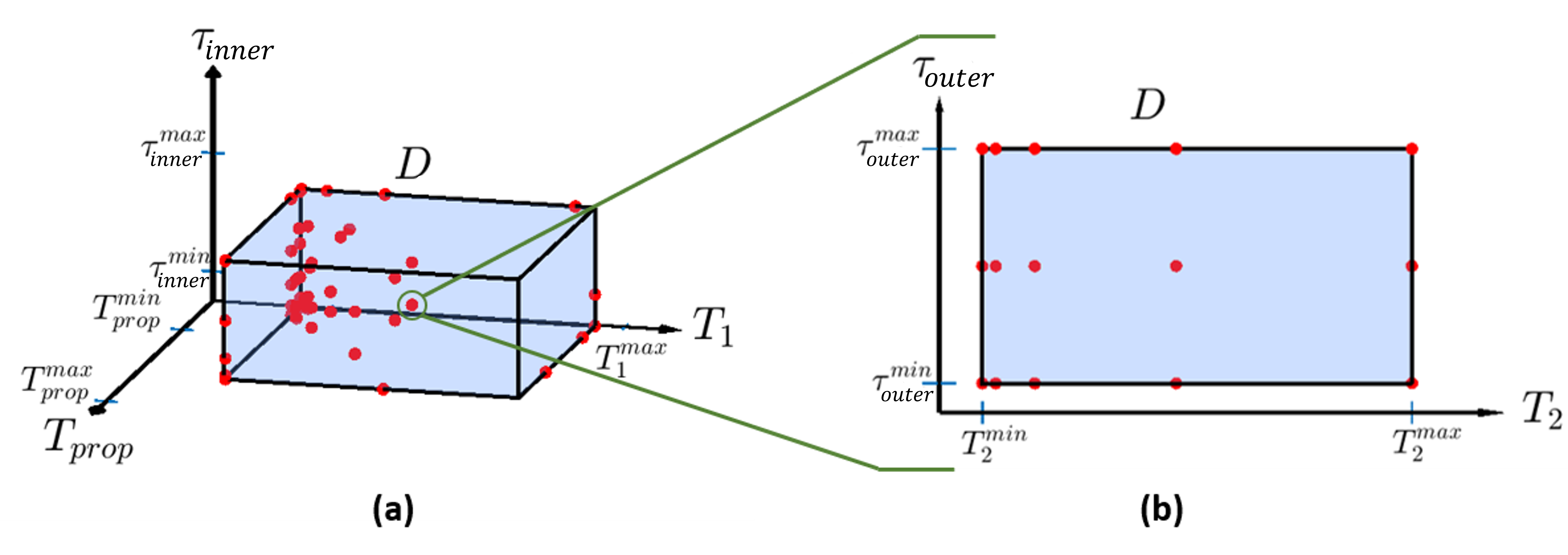} 
\end{center}
\caption{Illustration of discretized processes in the parameter space. (a) \(D_{att}\) and \(D_{alt}\) domains are discretized to obtain processes of \(\bar{D}_{att}\) and \(\bar{D}_{alt}\) shown by the red dots. (b) For each member process of \(\bar{D}_{att}\), a different set of \(\bar{D}_{lat}\) processes are obtained.}
\label{fig:full_discritization_process}
\end{figure}

\subsection{Identification Steps}
DNN-MRFT handles identification as a classification problem. Process classes in the unknown process parameters' space are  sampled based on the relative sensitivity metric and the minmax approach \cite{rohrer1965sensitivity} to ensure sampling efficiency, while ensuring full coverage. The relative sensitivity function indicates the robustness of the system to the changes in process parameters and is governed by the following equation:
\begin{equation}\label{eq_performance_deterioration}
J_{ij} = \frac{Q(C_i, G_j) - Q(C_j, G_j)}{Q(C_j, G_j)} \times 100 \%
\end{equation}
where \(J_{ij}\) represents the degradation in performance due to applying controller \(C_i\), which is the optimal controller for the process \(G_i\) and a sub-optimal controller for the process \(G_j\). \(Q\) denotes the integral square error (ISE) of the step response of the closed loop system. Note that \(J_{ij}\neq J_{ji}\) so we define \(J_{(ij)}=max(J_{ij},J_{ji})\).

We summarize the identification steps reported in \cite{Ayyad2020,ayyad2020multirotors} for the identification of the inner loop and outer loop dynamics as follows:
\begin{enumerate}
    \item Identify the domain of the unknown time parameters \(T_{prop}\), \(T_1\), \(T_2\), \(\tau_{in}\), and \(\tau_{out}\). We select these ranges based on prominent multirotor UAV sizes and designs (we include multirotor UAVs that span few centimeters to a few meters). Namely the selected parameters ranges are the same for those in \cite{ayyad2020multirotors} and they are \(T_{prop}\in[0.015,0.3]\), \(T_{1}\in[0.2,2]\), \(T_{2}\in[0.2,6]\), \(\tau_{in}\in[0.0005,0.03]\), and \(\tau_{out}\in[0.0005,0.15]\). Altitude parameters domain \(D_{alt}\), attitude parameters domain \(D_{att}\), and lateral motion parameters domain \(D_{lat}\) are formed based on the selected parameters' domains.
    
    \item Relative sensitivity value is selected to be \(J^*=J_{(ij)}=10\%\) which means that adjacent discretized processes would provide at worst 10\% drop in closed loop system performance.
    \item Discretize \(D_{att}\) and \(D_{alt}\) as illustrated in Fig. \ref{fig:full_discritization_process} (a) and outlined in \cite{Ayyad2020} based on \(J^*\) value to obtain discretized domains \(\bar{D}_{att}\) and \(\bar{D}_{alt}\).
    \item Discretize the outer loop domain \(D_{lat}\) for every discretized inner loop process in \(\bar{D}_{att}\) as shown in Fig. \ref{fig:full_discritization_process} (b). This will result in a set of discretized domains \(\{\bar{D}_{lat,1},...,\bar{D}_{lat,n}\}\) where \(n\) is the number of processes in \(\bar{D}_{att}\) \cite{ayyad2020multirotors}.
    
    \item In simulation, generate system responses for every process in \(\bar{D}_{att}\) and \(\bar{D}_{alt}\) based on \(\beta=-0.73\) value reported in \cite{Ayyad2020} and for \(\bar{D}_{lat,i}\) use \(\beta\) value that is specific for every process in \(\bar{D}_{att}\).
    
    \item Find the optimal controller for every discretized process in \(\bar{D}_{att}\), \(\bar{D}_{alt}\), and \(\bar{D}_{lat}\) using the tuning methodology presented in Section \ref{sec:ControllerTuning}.
    
    \item Train the DNN based on simulation data. Generated data is augmented with noise and process bias for better generalization. The output layer uses the modified softmax cost function derived in \cite{Ayyad2020}.
    
    \item Finally, excite a periodic motion experimentally using MRFT algorithm shown in Equation (\ref{eq_mrft_algorithm}) and feed the measured process output to the DNN for identification and obtain new controller parameters.
    
\end{enumerate}

\section{Controller Design}\label{ControllerDesign}

The proposed control design approach is built around two facts. First, the obtained equivalent closed-loop linear system representation given in Eqs. \eqref{eq_closed_loop_alt} and \eqref{eq_closed_loop_lat} permits the use of a handful of linear feedback control design tools. Second, the close match between the obtained simulation models and experimentation, as illustrated by the results in Section \ref{sec:results}, alleviates the need for model and controller refinement through repetitive experimentation.

\subsection{Controller Structure Selection}
We could use either PID or PD control structure for the outer and the inner control loops. The selection depends on a few factors, and a single control structure cannot be the best for all scenarios (e.g. external disturbance, reference signal type, robustness, etc.). 

It was found, based on the extensive simulations where different permutations of controller structures and tuning methods were used, that fast inner dynamics are essential for high performance trajectory tracking. Hence, a PD control structure for the inner loops was selected over PID as it provides lower rising time \(T_r\). The inner loop PID controller would not suffer from steady-state errors due to asymmetric CoG or motor thrusts imbalance. This necessitates the use of a PID structure in the outer loop if a PD structure is used in the inner loop and the steady-state error needs to be eliminated. The results of steady-state analysis is therefore limited to the case when a PD controller is used in the inner loops, and a PD or a PID controller is used in the outer loops. The overall controller structure is shown in Figure \ref{fig:controller_struct}. The steady-state errors for step and ramp excitations applied as reference signals or force disturbances (e.g. due to wind) are analyzed. We choose to analyze ramp excitations to evaluate steady-state behavior of the system when the reference is varying (i.e. trajectory) or when the source of disturbance is ramped (e.g. approaching wind source). The results of the steady-state error analysis are summarized in Table (\ref{table:ess}).

It can be seen that the \(\text{PD}_{in}/\text{PID}_{out}\) configuration is a suitable choice for eliminating steady-state error for both varying references and constant disturbances cases. This means that the tuning of the outer PID controller will be based on a ramp input reference. We will also consider the \(\text{PD}_{in}/\text{PD}_{out}\) configuration tuned for step input based on its high performance and consistency seen in practice. The \(\text{PD}_{in}/\text{PD}_{out}\) configuration suffers steady-state errors except for a step reference input, but its performance can be improved by using the higher order references as discussed in section \ref{TrajGen}.

will be used with the following simplifications:
\begin{itemize}
    \item Feedback linearization will be ignored, obviously due to linearization around nominal operating point.
    \item Velocity and acceleration reference inputs are set to zeros.
\end{itemize}

\begin{table*}
\centering
\begin{tabular}{||c|c|c|c|c||}
\hline
{Controller Configuration} & \multicolumn{2}{c|}{\text{$e_{ss}$ due to reference}} & \multicolumn{2}{c||}{\text{$e_{ss}$ due to disturbance}} \\ \cline{2-5} 
 & Step & Ramp & Step & Ramp \\ \hline
\(\text{PD}_{in}/\text{PD}_{out}\) & 0 & $\dfrac{1}{1+K_{p}K_{out}}$ & $-\dfrac{K_{out}}{1+K_{p}K_{out}}$ & $-\infty$ \\ \hline
\(\text{PD}_{in}/\text{PID}_{out}\) & 0 & 0 & 0 & $-\dfrac{1}{K_{i}}$ \\ \hline
\end{tabular}
\caption{Steady-state errors for different inputs references and disturbances.}
\label{table:ess}
\end{table*}

\subsection{Controller Tuning}\label{sec:ControllerTuning}
The tuning is mainly based on minimizing a cost function \(Q\), which is a function of the error to a specific input to the system. In our work we will choose $Q$ the ISE performance index realized by:
\begin{equation}\label{eqn:ISE}
    Q_{ISE} = \int_0^t ({}^Ip_x^{ref}(t) - {}^Ip_x(t))^2 dt 
\end{equation}

Other performance indices were investigated, such as Integral Absolute Error (IAE), Integral Time-weighted Absolute Error (ITAE) and Integral Time-weighted Squarded Error (ITSE). It is not convenient to tune controllers in the presence of stead-state errors, as the value of the cost functional is not bounded when \(t\rightarrow \infty\). This is due to the introduction of some additional optimization parameters like the simulation time \(T\), as we are tuning based on simulated system response. Therefore, we exclude tuning controller structures with excitation sources that would result in steady-state errors (refer to Table \ref{table:ess}.

\begin{figure}
    \centering
    \includegraphics[width=\linewidth]{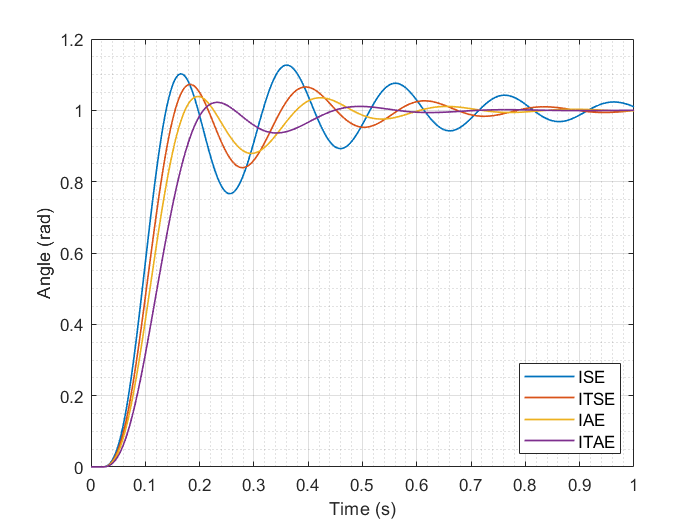}
        \caption{Step response of \(PD_{in}\) controller tuned based on different criteria for a step input.}
        \label{fig:InnerLoopDiffCriPD}
\end{figure}

\begin{table}
\centering
\begin{tabular}{||c|c|c|c|c|c|c|c||}
\hline
Tuning Criteria & \(T_r\) (s) & PO (\%) & \(T_s\) (s) \\ \hline
IAE    & 0.0897 & 3.6458  & 0.4577 \\ \hline
ISE    & 0.0692 & 10.7819 & 0.8926 \\ \hline
ITAE   & 0.1077 & 2.5773  & 0.2915  \\ \hline
ITSE   & 0.0796 & 6.9892  & 0.5308  \\ \hline
\end{tabular}
\caption{Step response characteristics of \(PD_{in}\) controller tuned based on different criteria for a step input.}
\label{table:PD_inner_step_char}
\end{table}

The tuning is conducted  using MATLAB and Simul\-ink. We used the derivative free Nelder-Mead simplex algorithm with inequility constraints is realized by the \textit{"fminsearchbnd"} optimization function \cite{fminsearchbnd2021}. The objective of the optimization function is to minimize the error function, and the decision variables are the controller parameters. We found that the optimizer usually get stuck at local minima when tuning PID controller parameters in their parallel form. We have also found that parameterizing the control parameters using the \textit{homogeneous tuning rules} avoided the local minima issue, and resulted in a much faster optimization time. The equation for the homogeneous tuning rules are given by \cite{boiko_non-parametric_2012}:
\begin{equation}\label{eqn:nonparamtunrul}
    K_c = c_1\frac{1}{|G(j\Omega_0)|} \; , \; T_I = c_2\frac{2\pi}{\Omega_0} \; , \; T_D = c_3\frac{2\pi}{\Omega_0}
\end{equation}
Here $c_1, c_2, c_3$ are constant parameters which define the homogeneous tuning rule, \(\Omega_0\) is the frequency response of the system at test phase \(\beta\), and \(|G(j\Omega_0)|\) is the amplitude response of the system at the same test phase \(\beta\) (refer to Equation \eqref{eq_mrft_algorithm}). If we choose to tune only a PD controller, then \(c_2\), the constant corresponding to the integral term, will be equal to zero. The homogeneous tuning rule is related to the gain and phase margins of the system by two sets of equations \cite{Boiko2013}. These equations govern a relation between the homogeneous tuning rule parameters and allow us to exploit the time and gain invariance properties of the tuning rule to impose inequality constraints on the gain and phase margins. The phase margin is related to the homogeneous tuning rule parameters by:
\begin{equation}\label{eqn:PMconstraints}
\begin{split}
    \beta = \sin\left(\varphi_m + arctan\left(\frac{1}{2\pi c_2} - 2\pi c_3\right)\right) , & \\    c_1 \sqrt{1+\left(2\pi c_3 - \frac{1}{2\pi c_2}\right)^2} = 1 , &  
\end{split}
\end{equation}
where \(\varphi_m\) is the phase margin of the closed loop system. On the other hand, the gain margin constrain equations are given by:
\begin{equation}\label{eqn:GMconstraints}
\begin{split}
    \beta = -\dfrac{2\pi c_3 - \dfrac{1}{2\pi c_2}}{\sqrt{1+\left(2\pi c_3 - \dfrac{1}{2\pi c_2}\right)^2}} , & \\   \gamma_m c_1 \sqrt{1+\left(2\pi c_3 - \dfrac{1}{2\pi c_2}\right)^2} = 1 . &   
\end{split}
\end{equation}
where \(\gamma_m\) is the gain margin of the closed loop system. If we want to impose phase margin constraints, we use Equation \eqref{eqn:PMconstraints} with Equation \eqref{eqn:nonparamtunrul} to find PID parameters. Similarly, we use Equation \eqref{eqn:GMconstraints} with Equation \eqref{eqn:nonparamtunrul} to find PID parameters with gain margin constraints. Apparently, using either of the constraint sets whould result in the same PID tuning if no constraints on the stability margins are imposed.

For example, in a gain margin based tuning, we optimize \(\gamma_m\), \(c_1\), and \(c_3\) and use Equation \eqref{eqn:GMconstraints} to calculate the other parameters directly (i.e. \(c_2\) and \(\beta\)). The parameter \(\beta\) is a phase parameter that we can use to find the corresponding frequency and amplitude responses of the open loop system \(G(j\Omega)\) being optimized (i.e. \(\Omega_0\) and \(|G(j\Omega_0)|\). The tuning rule in Equation \eqref{eqn:nonparamtunrul} is then fully defined, and it is possible to calculate the PID parameters. Because \(\gamma_m\) is a decision variable, we can impose a minimum gain margin for the system, which results in robust tuning.

We have tried the above tuning procedure for the different controller structures that we might use. We found that tuning a \(\text{PD}_{in}/\text{PD}_{out}\) structure against a step reference excitation resulted in the same controller parameters of a \(\text{PD}_{in}/\text{PID}_{out}\) structure (i.e. the I-term of the outer PID is set to zero by the optimizer). Therefore, we excluded the \(\text{PD}_{in}/\text{PID}_{out}\) structure with step reference excitation signal from our tuning options. Also, we found that tuning a \(\text{PD}_{in}/\text{PID}_{out}\) structure with ramp reference resulted in tuning parameters that are insensitive to the application of step disturbances. Therefore, we simply tuned a \(\text{PD}_{in}/\text{PID}_{out}\) structure with ramp reference without applying any disturbance signal. Overall, from all permutations of controller structures and excitation signals we settled on two sets of controller structures and excitation signals to optimize for: a \(\text{PD}_{in}/\text{PD}_{out}\) structure with step reference, and  \(\text{PD}_{in}/\text{PID}_{out}\) structure with ramp reference. For convenience, we refer to those as \(\text{PD}_{in}/\text{PD}_{out}\) and \(\text{PD}_{in}/\text{PID}_{out}\) structures or configurations. We found that the \(\text{PD}_{in}/\text{PID}_{out}\) configuration resulted in large oscillations when performing hover or step following. Thus the \(\text{PD}_{in}/\text{PD}_{out}\) configuration is always used to perform hover and step following tasks.

Settling on the ISE criterion over the others for inner and outer loop tuning was based on the speed of the response. Figure \ref{fig:InnerLoopDiffCriPD} and Table \ref{table:PD_inner_step_char} shows the step responses characteristic of inner loop PD controllers tuned based on the four criteria mentioned for a step input. It is obvious that ISE tuned controller resulted in the lowest rise time \(T_r\) among all other criteria. As the speed of the inner loop controller has a great impact on how well the outer loop controller performs, we chose ISE as a criterion for inner loop tuning. Similarly, we used ISE for the outer loop tuning of \(\text{PD}_{in}/\text{PD}_{out}\) and \(\text{PD}_{in}/\text{PID}_{out}\) configurations as it provided the least rising time \(T_r\) among all other criteria at the trade-off of providing oscillatory response. Having a quick system response is important in the reduction of temporal tracking errors when performing fast trajectories.

\begin{figure}
    \centering
    \includegraphics[width=\linewidth]{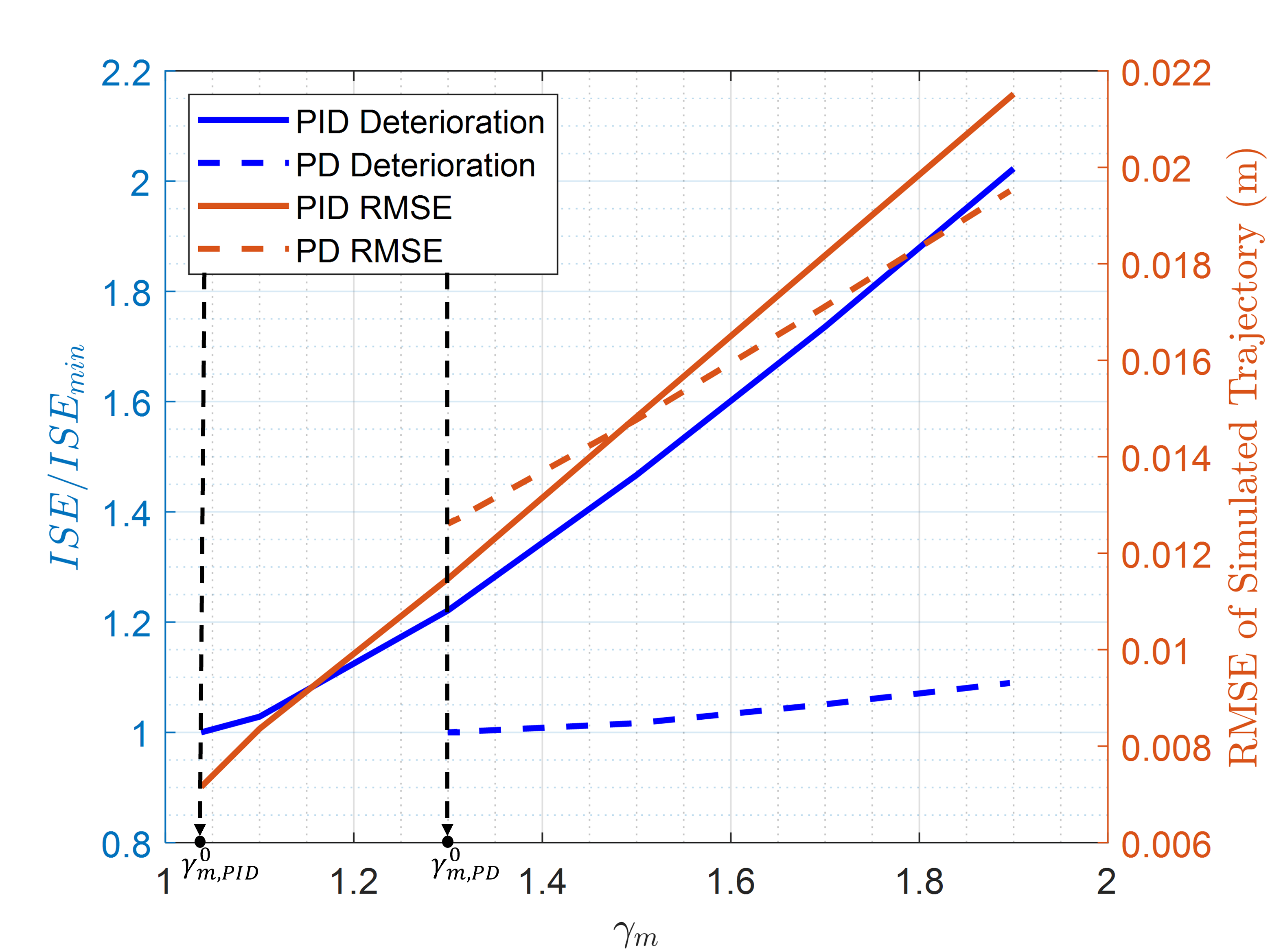}
        \caption{Performance robustness trade-off curve for PID controller with gain margin constraint. The simulated trajectory is a \(3\times1.5m\) figure-eight.}
        \label{fig:PerfGM}
\end{figure}

The simulation results for the performance-robustness trade-off curve of the \(\text{PD}_{in}/\text{PD}_{out}\) and \(\text{PD}_{in}/\text{PID}_{out}\) configurations when gain margin based tuning is used are shown in Figure \ref{fig:PerfGM}. The actual numerical results are based on the experimental quadrotor we used in this paper (without payload). Still, based on extensive simulations we found that the qualitative behaviour is applicable for almost all realizable multirotor UAV designs. We chose to have gain margin based tuning as variations in the drag mainly alter the system gain. The \(\text{PD}_{in}/\text{PID}_{out}\) configuration optimal tuning (i.e. unconstrained) resulted in a gain margin of \(\gamma_{m,PID}^0 = 1.038\), which is close to the instability limit. The \(\text{PD}_{in}/\text{PID}_{out}\) configuration is recommended for figure-eight trajectory tracking task when the best performance is required. But the \(\text{PD}_{in}/\text{PD}_{out}\) configuration results in a flat deterioration performance curves, providing better performance choice when larger stability margins are required.

\section{Results}
\label{sec:results}
In this section, we will show a set of trajectory tracking experiments with different controller tuning settings of the \(\text{PD}_{in}/\text{PD}_{out}\) and \(\text{PD}_{in}/\text{PID}_{out}\) configurations. First, we present the adaptation capability to system dynamics changes, then we show the robust tuning effect and the behaviour when the system is subjected to system dynamics changes or external disturbances. We conclude by comparing our results with the literature where it is shown that the results obtained in this work are the state-of-the-art for high speed trajectory tracking under external wind disturbance. A video that summarizes the experimental results can be found in \cite{paper_video}.

\subsection{Experimental Setup}
The quadrotor platform used in the experiments of this paper is the QUANSER QDrone. It has a carbon fiber frame with a protected propulsion system design that consists of four Cobra CM-2205/2100kv BLDCs and 6045 polycarbonate propellers. The QDrone weighs 1125 g with a thrust-to-weight ratio of $1.9$. QDrone features Intel Aero Compute on-board embedded computer with the ability to be programmed through MATLAB/Simulink. The Madgwick filter \cite{madgwick2011} is used to estimate the attitude of the multirotor through the on-board BMI160 IMU Sensor. OptiTrack motion capture system was used to provide position and yaw measurements to the multirotor at 250 Hz.

\subsection{Performance Assessment}
To assess the trajectory tracking performance, we use two metrics, the first one is the root mean squared error (RMSE) which indicates how good the temporal tracking is:

\begin{equation}
    RMSE = \sqrt{\dfrac{\sum _{i=1} ^N ({}^Ip_x-{}^Ip_x^{ref})_i^2 + ({}^Ip_y-{}^Ip_y^{ref})_i^2}{N}}
\end{equation}

where $N$ is the number of the sampled data points in the trajectory. The second metric is the contouring error, which captures how good the spatial tracking is, and is defined by the minimum Euclidean distance between a given position measurement and the nearest point on the reference trajectory:
\begin{equation}
    CE_i = \min \left( \sqrt{({}^Ip_{x_i} - {}^Ip_x^{ref})^2 + ({}^Ip_{y_i} - {}^Ip_y^{ref})^2} \right)
\end{equation}
It follows that we define the average of all summed contouring errors to be \(CE_{avg}\), and the maximum contouring error to be \(CE_{max} = \arg\!\max CE_i\).

In our flight experiments, we use figure-eight trajectory with dimensions of \(3\times1.5m\) which is generated as described in Section \ref{TrajGen}, without imposing any actuator limits in the trajectory optimization phase. We have then used The fastest trajectory we were able to achieve before reaching motors' upper limits took eight seconds to perform. We have also used a smaller \(1.5\times0.75m\) figure-eight trajectory in the Y-Z plane to be compared with the performance shown in \cite{o2021meta} for tracking the same trajectory under wind disturbance. In all flight experiments, we will compare the results with a nonlinear simulation model utilizing the parameters acquired from the DNN-MRFT identification.

\begin{figure}[]
    \includegraphics[width=\linewidth]{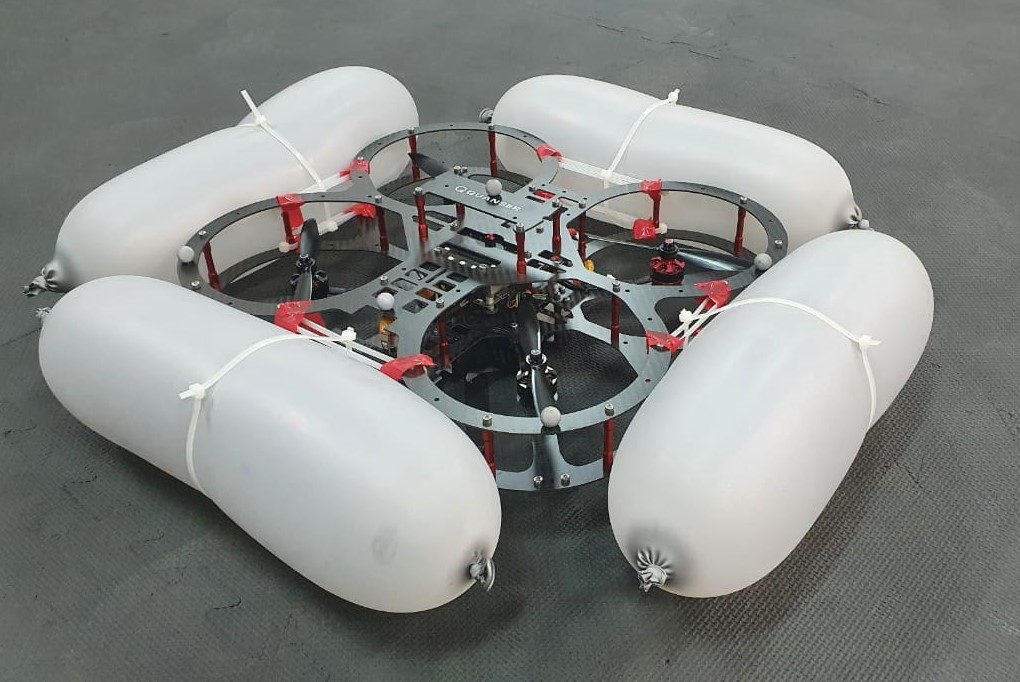}
        \caption{The QDrone with the dynamics changing payload installed.}
        \label{fig:QDroneWPayload}
\end{figure}

\subsection{Online Adaptation to Payload (Physical) Changes and Robustness}\label{OnAdaptCap}
We demonestrate the in-flight adaptability behaviour using the \(\text{PD}_{in}/\text{PD}_{out}\) configuration. The same is applicable for the \(\text{PD}_{in}/\text{PID}_{out}\) configuration or any other possible tuning configuration.

\begin{figure*}
    \begin{center}
    \includegraphics[width=\textwidth]{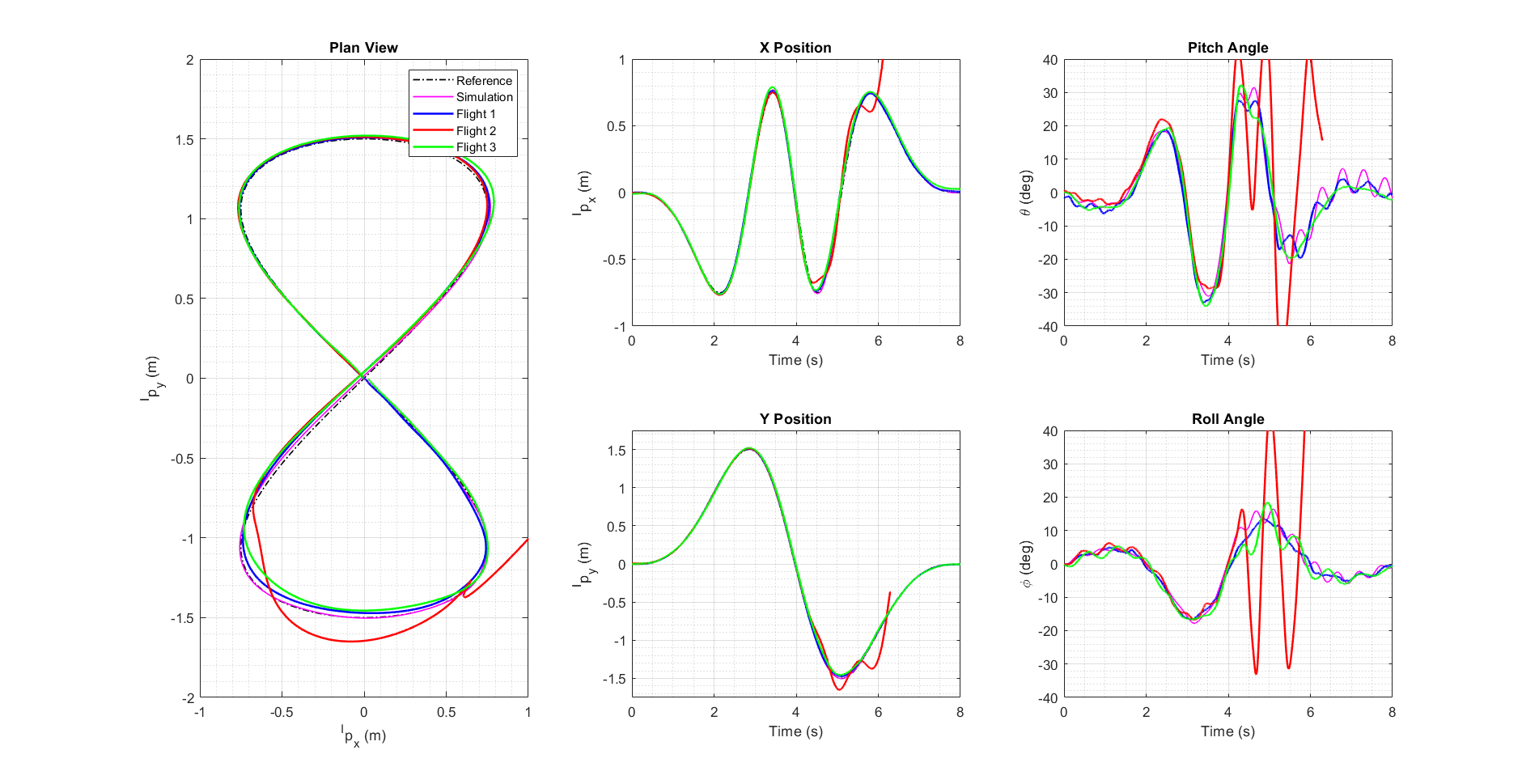}
    \caption{Tracking performance of the three flights with PD as outer loop controller. Flight 1 (blue): PD Gains Set I - Without Payload. Flight 2 (red): PD Gains Set I - With Payload, QDrone crashed resulting in discontinuation of data streaming. Flight 3 (green): PD Gains Set II - With Payload. }
    \label{fig:3casesfig_time_ang}
    \end{center}
\end{figure*}

The \(\text{PD}_{in}/\text{PD}_{out}\) configuration gains were tuned without imposing any robustness margins (i.e. resulting in \(\gamma_{m,PD}^0=1.30\) gain margin shown in Figure \ref{fig:PerfGM}). These gains were tuned based on DNN-MRFT identified system parameters. Three trajectory tracking flights were done in sequence. In the first flight, we used the \(\text{PD}_{in}/\text{PD}_{out}\) configuration tuned based on DNN-MRFT applied to the multirotor UAV without any payload attached (we refer to it as the Gains Set I). The achieved tracking errors were as low as $2.6 \; cm$ which is on par with the state-of-the-art in trajectory tracking performance for multirotor UAVs in the given range of velocities and platform sizes. In fact, the achieived performance exceeds the performance reported in the recent work of \cite{Faessler_2018}, where an error of $3.3 \; cm$ was achieved, approximated from a \(4\times2m\) figure-eight trajectory with a similar maximum speed of around $3.2 \; m/s$ , and using a platform that has a thrust-to-weight ratio of 4. Before the second flight and while the multirotor UAV is still in hover, we attach a payload to it to alter its aerodynamic properties. The light-weight payload consists of four $12 \; cm$ wide by $40 \; cm$ long balloons installed on each side of the UAV as shown in figure \ref{fig:QDroneWPayload}. While performing the second flight with the Gains Set I, the platform exhibited instability and it crashed, apparently due to the altered dynamics. Before the third flight, we re-run DNN-MRFT identification and tuning; which is performed online and takes just a few seconds to adapt for the newly added payload. Table \ref{table:MRFT_OAC_results} shows the DNN-MRFT identification results for both the payload and the no payload cases. The achieved trajectory tracking RMSE error with the new controller tuning (Gains Set II) on the platform with added payload is $3.42 \; cm$, which is slightly higher than the reported errors for the first flight. Table \ref{table:OACexps} summarizes the performance results for the online adaptation experiments (Flights 1, 2 and 3). Figure \ref{fig:3casesfig_time_ang} shows the measured position and angles for the three flights performed. 

The instability that occurred in the second flight when Gains Set I was used can be explained by simulating the identification results. Applying the Gains Set I to the identified lateral dynamics of the multirotor UAV with payload (i.e. flight scenario 2) resulted in a phase margin of \(\varphi_m = 3.45^o\) and a gain margin of \( \gamma_m = 1.12\). With such small stability margins it is likely that the nonlinearities in the system would alter the dynamics in a way that would result in instability. On the other hand, the stability margins in the first and third flight scenarios were around $\varphi_m = 10^o ,\; \gamma_m = 1.3$ for the same lateral dynamics.

We also demonstrate robust tuning of controllers for payload changes. We demonstrate such robustness using the \(\text{PD}_{in}/\text{PID}_{out}\) configuration. Because the unconstrained tuning of the \(\text{PD}_{in}/\text{PID}_{out}\) configuration results in a low gain margin of \(\gamma_{m,PID}^0 = 1.038\), we chose to introduce a minimum gain margin constraint of \(\gamma_m = 1.5\) at the cost of having an increased ISE cost by 46.78\% compared to the minimum for the sake of increased robustness. By simulating the figure-eight trajectory it is also anticipated that the RMSE would increase by 107.3\% due to the sub-optimality in the tuning.

\begin{figure*}
    \centering
    \includegraphics[width=\textwidth]{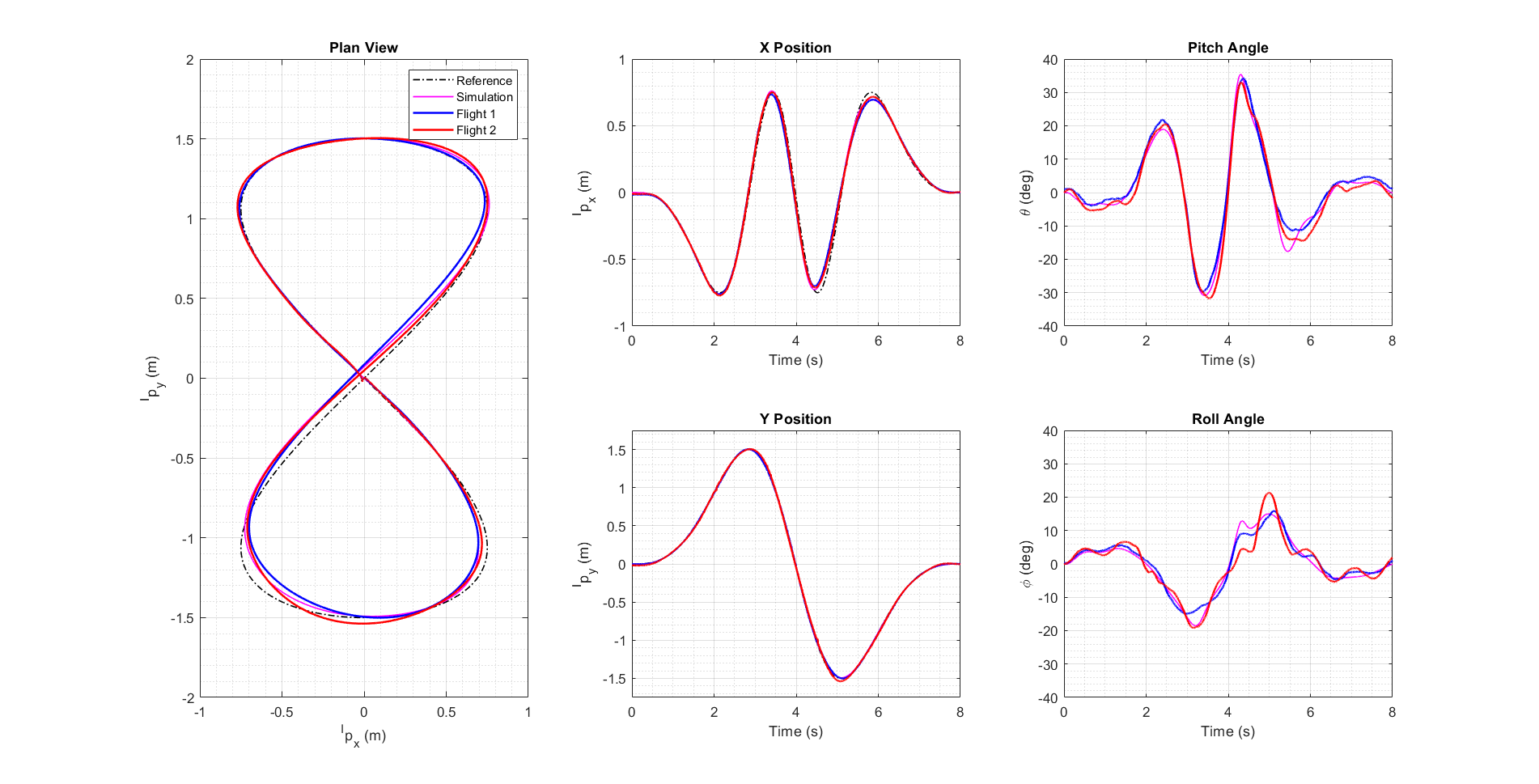}
    \captionsetup{justification=centering}
        \caption{Tracking performance of the two PID scenarios (robust tuning). Flight 4 (blue): PID Gains Set I - Without Payload. Flight 5 (red): PID Gains Set I - With Payload. The qualitative match between simulation response and experimental results is clear in the angle responses.}
        
        \label{fig:PID2cases}
\end{figure*}

To test the robustness of the proposed controller we have done two consecutive flight tests with the robust PID tuning: the first without the payload, and the second with the payload installed. Results for the two flights can be seen in Figure \ref{fig:PID2cases} while the reported performance figures can be found in Table \ref{table:OACexps} (Flights 4 and 5). The results obtained for \(\text{PD}_{in}/\text{PD}_{out}\) non-robust tuning showed better performance compared to the \(\text{PD}_{in}/\text{PID}_{out}\) robust tuning. Also the performance of the \(\text{PD}_{in}/\text{PID}_{out}\) configuration got slightly improved when the payload got installed due to the increase in the systems gain as seen in the DNN-MRFT identification results in Table \ref{table:MRFT_OAC_results}, thus sacrificing some of the gain margin that the controller was tuned on. These results confirm the suitability of the suggested tuning framework on providing powerful design tools that can be used to balance performance versus robustness.

The qualitative match between the simulation responses and the experimantal results is clear for the responses of the angles shown in Figure \ref{fig:PID2cases}. This is an indication of the suitability of the suggested model structure we used to design the attitude and lateral dynamics controllers.

\begin{table*}[!h]
\centering
\begin{tabular}{||c|c|c|c|c|c|c|c||}
\hline
                & $K_{in}$ & $K_{out}$ & $T_{prop}$ & $T_1$  & $T_2$  & $\tau_{in}$ & $\tau_{out}$ \\ \hline
Without Payload & 68       & 1.1516     & 0.064      & 0.2494 & 1.1629 & 0.0009      & 0.0005       \\ \hline
With Payload    & 37.46    & 5.01       & 0.0498     & 0.1938 & 6      & 0.014       & 0.0005       \\ \hline
\end{tabular}

\caption{DNN-MRFT identification results for the QDrone with and without the payload for the Y and Roll control loops.}
\label{table:MRFT_OAC_results}
\end{table*}

\begin{table}[!h]
\begin{tabular}{||c|c|c|c||}
\hline
         & $RMSE (cm)$ & $CE_{avg} (cm)$ & $CE_{max} (cm)$ \\ \hline
Flight 1 & 2.6         & 1.07            & 3.88      \\\hline
Flight 2 & Unstable     &  Unstable     &    Unstable \\\hline 
Flight 3 & 3.42        & 1.76            & 8.01 \\\hline

Flight 4 & 4.22         & 1.81            & 10.4            \\\hline
Flight 5 & 3.33        & 1.53            & 7.64 \\\hline
Flight 6 & 3.78         & 1.99            & 7.33            \\\hline
Flight 7 & 3.59        & 1.47            & 5.26 \\\hline
\end{tabular}  

\caption{Performance results summary for all flights.}
    \label{table:OACexps}
\end{table}

\begin{figure*}
    \centering
    \includegraphics[width=\linewidth]{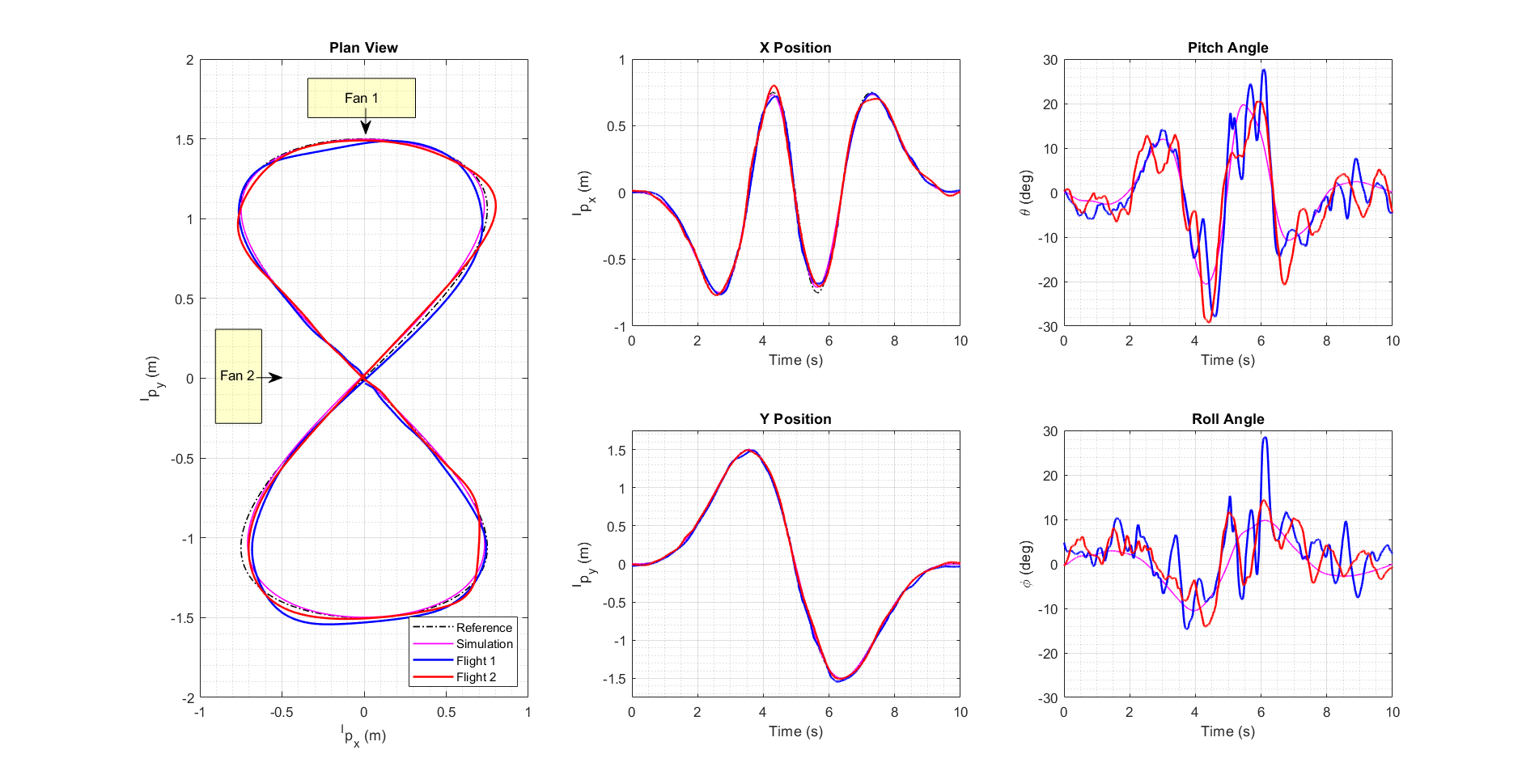}
        \caption{Tracking performance under wind disturbance. Fans locations indicated by the yellow boxes. Flight 6 (blue): PD under wind disturbance. Flight 7 (red): PID under wind disturbance. The simulation results shown are obtained without disturbance.}
        \label{fig:disturbed8sh}
\end{figure*}

\subsection{Tuning for Wind Disturbance Attenuation}

We performed two flight tests to assess the performance of the \(\text{PD}_{in}/\text{PD}_{out}\) and \(\text{PD}_{in}/\text{PID}_{out}\) configurations under external wind disturbances. The \(\text{PD}_{in}/\text{PD}_{out}\) and \(\text{PD}_{in}/\text{PID}_{out}\) controllers were tuned as in the previous experiment; without imposing any gain margin and with a gain margin of \(\gamma_m=1.5\) respectively. 

As found in the stead-state analysis presented in Table \ref{table:ess} and confirmed by experimentation, the \(\text{PD}_{in}/\text{PD}_{out}\) configuration resulted in a constant bias both during hover and also when tracking a trajectory under external wind disturbances, which resulted in high RMSE and \(CE_{avg}\) errors. The fact that the amount of the steady-state error observed during hover and trajectory tracking is the same allowed us to compensate for the steady-state error when at hover. The same figure-eight trajectory was tracked without the payload for ten seconds under high speed wind reaching $5 \; m/s$ applied laterally from two wind sources. We could not fly the trajectory with a lap speed of eight seconds as in the previous experiment, as the motors reached saturation when counteracting the wind disturbance. The tracking performance under wind disturbances seen in Figure \ref{fig:disturbed8sh} is comparable with the wind-free case. Table \ref{table:OACexps} (Flights 6 and 7) summarizes the performance results for the figure-eight trajectory tracking under wind disturbance experiments. The \(\text{PD}_{in}/\text{PID}_{out}\) configuration performed better despite having stricter gain margins, and it also resulted in noticeably smaller variations in the tilt angles indicating smoother controller actions compared to the \(\text{PD}_{in}/\text{PD}_{out}\) configuration.

\begin{figure}[]
    \centering
    \includegraphics[width=\linewidth]{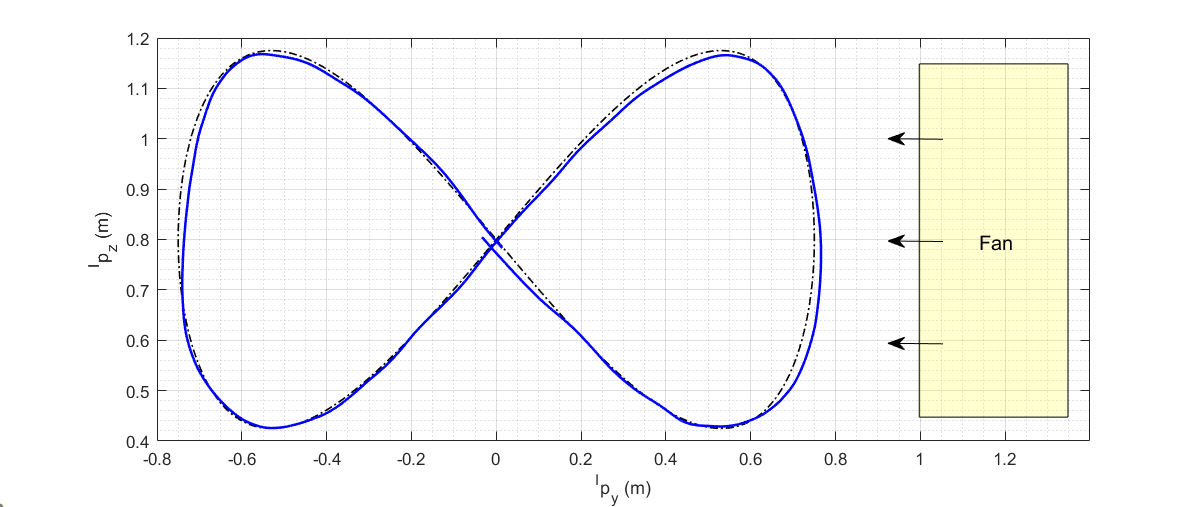}
        \caption{Tracking performance for a vertical (Y-Z plane) \(1.5\times0.75m\) figure-eight trajectory under wind disturbance.}
        \label{fig:Vfig8}
\end{figure}

We also replicated a vertical figure-eight trajectory flight under wind disturbances to further assess the disturbance attenuation capabilities of our tuned controller and to replicate the trajectory tracking results reported in \cite{o2021meta}. The trajectory flown is a \(1.5\times0.75m\) figure-eight trajectory in the Y-Z plane and lasts for 8 seconds under wind speeds of \(5 m/s\). The \(\text{PD}_{in}/\text{PID}_{out}\) configuration was used in this experiment. Our flight experiments show drastic improvements in the performance with an error of \(1.3 \; cm\) compared to \(20 \; cm\) in \cite{o2021meta}. The results of this experiment can be shown in Figure (\ref{fig:Vfig8}) and Table \ref{table:V8figWindExp}.

\begin{table}[h!]
\centering
\begin{tabular}{||c|c||}
\hline
         & $RMSE (cm)$  \\ \hline
Our Work & 1.3             \\\hline
O'Connell et. al \cite{o2021meta} & 20        \\\hline           
\end{tabular}
\caption{Performance comparison of a vertical (Y-Z plane) \(1.5\times0.75m\) figure-eight trajectory under wind disturbance.}
        \label{table:V8figWindExp}
\end{table}

\section{Conclusion}
A controller tuning approach based on DNN-MRFT identification to perform accurate trajectory tracking was presented. The suggested approach provides several demonstrated advantages over existing methods in the literature. First, controller parameters' tuning can be done systematically based on the real-time identification results of DNN-MRFT. This capability was demonstrated experimentally, where the multirotor UAV was able to adapt in-flight to significant changes in the payload that could alter the stability of the multirotor UAV. The second advantage is the ability to trade-off performance and robustness, which was demonstrated by the tuning of a robust PID controller that performed well despite payload changes and external wind. The third demonstrated advantage is the close match between simulation and experimentation which can be used to realize efficient planning algorithms, fault detection, etc. The in-flight tuned multirotor UAV achieved state-of-the-art performance in tracking a figure-eight trajectory.

A potential continuation of this work is to design a gain scheduling mechanism for improved control performance based on changing flight conditions. In future work, we aim to utilize high fidelity simulations to design real-time planning algorithms that could better leverage the UAV capabilities. We have also observed some interesting limit cycle behavior when using MRFT with feedback linearization. These limit cycles worth further investigations as they could be utilized for better identification of fast actuator and sensor dynamics.

\section*{Declaration of competing interest}

The authors declare that they have no known competing financial interests or personal relationships that could have appeared to influence the work reported in this paper.

\section*{Acknowledgment}
This work was supported by Khalifa University Grants CIRA-2020-082 and RC1-2018-KUCARS. We would like to thank Quanser team for their generous and timely support. We also thank Prof. Igor Boiko for the fruitful technical discussions. We also thank Eng. Mohammad Wahbah and Eng. Oussama Abdul Hay for helping in the preparation of the experimental setup.





\bibliographystyle{unsrt}
\bibliography{main}

\newpage
\newpage

\begin{IEEEbiography}[{\includegraphics[width=1in,height=1.25in,clip,keepaspectratio]{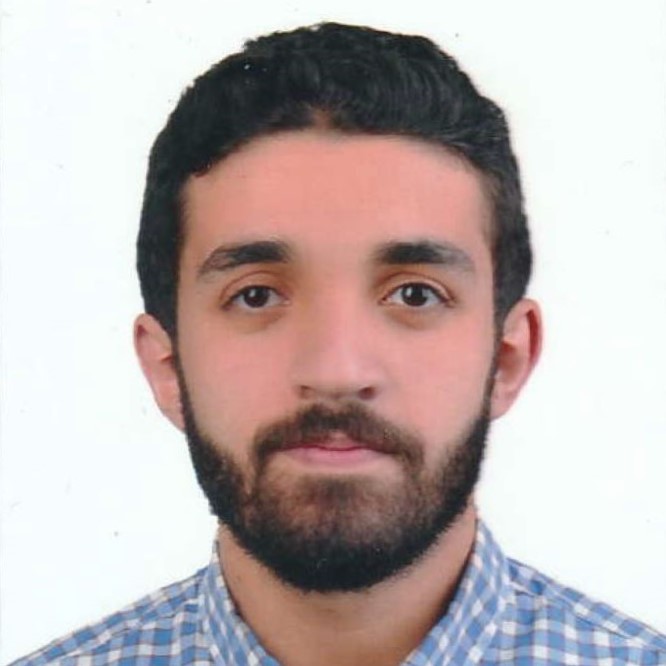}}]{AbdulAziz Y. AlKayas} received his BSc. degree in Electromechanical Engineering from Alexandria University, Alexandria, Egypt, in 2019, and his MSc. degree in Mechanical Engineering from Khalifa University, Abu Dhabi, UAE, in 2021. He is currently pursuing a PhD in Engineering degree with specialization in Mechanical Engineering in Khalifa University. He is mainly interested in research topics related to control of dynamical systems, autonomous robotics, as well as the design of aerial, marine and space systems.

\end{IEEEbiography}

\begin{IEEEbiography}[{\includegraphics[width=1in,height=1.25in,clip,keepaspectratio]{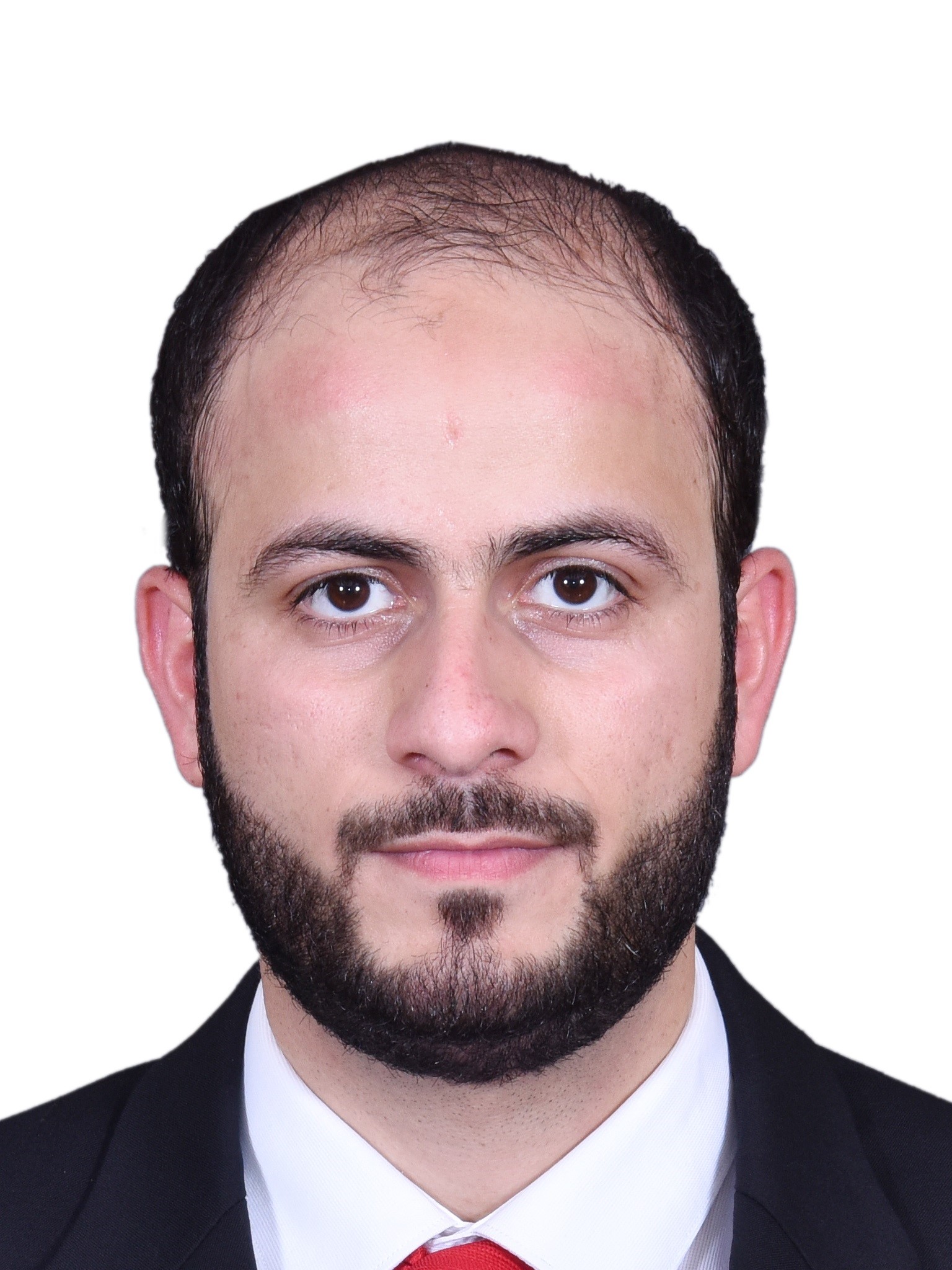}}]{Mohamad Chehadeh} received his MSc. in Electrical Engineering from Khalifa University, Abu Dhabi, UAE, in 2017. He is currently with Khalifa University Center for Autonomous Robotic Systems (KUCARS). His research interest is mainly focused on identification, perception, and control of complex dynamical systems utilizing the recent advancements in the field of AI.

\end{IEEEbiography}

\begin{IEEEbiography}[{\includegraphics[width=1in,height=1.25in,clip,keepaspectratio]{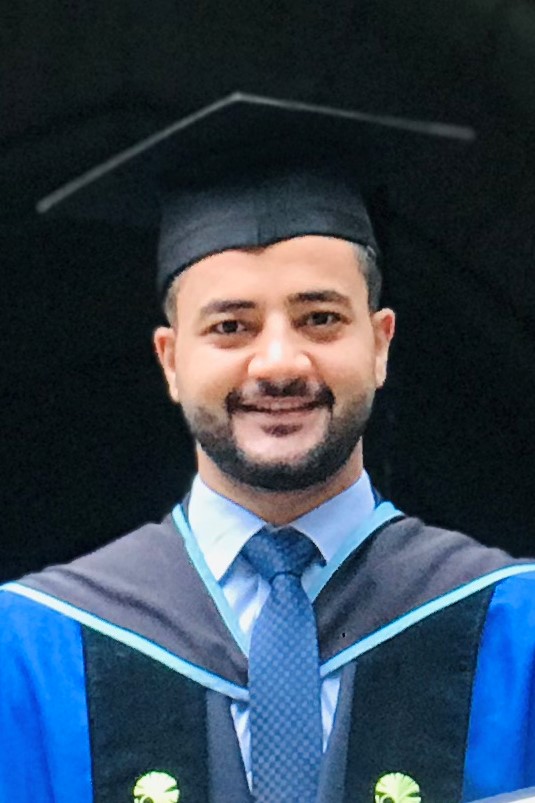}}]{Abdulla Ayyad} received his MSc. in Electrical Engineering from The University of Tokyo in 2019 where he conducted research in the Spacecraft Control and Robotics laboratory. He is currently a Research Associate in Khalifa University Center for Autonomous Robotic Systems (KUCARS) and the Aerospace Research and Innovation Center (ARIC) working on several robot autonomy projects. His current research targets the application of AI in the fields of perception, navigation, and control.
\end{IEEEbiography}

\begin{IEEEbiography}[{\includegraphics[width=1in,height=1.25in,clip,keepaspectratio]{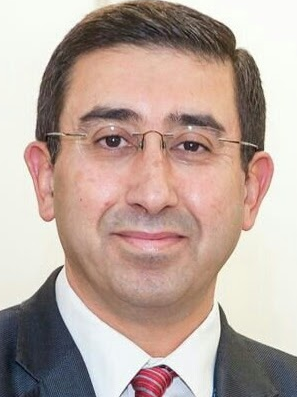}}]{Yahya Zweiri} received the Ph.D. degree from the King’s College London in 2003. He is currently an Associate Professor with the Department of Aerospace, Khalifa University, United Arab Emirates. He was involved in defense and security research projects in the last 20 years at the Defence Science and Technology Laboratory, King’s College London, and the King Abdullah II Design and Development Bureau, Jordan. His central research focus is interaction dynamics between unmanned systems and unknown environments by means of deep learning, machine intelligence, constrained optimization, and advanced control. He has published over 100 refereed journal and conference papers and filed ten patents in USA and U.K. in unmanned systems field.
\end{IEEEbiography}

\EOD

\end{document}